\documentclass[twoside,11pt]{article}

\pdfoutput=1
\usepackage[T1]{fontenc}
\usepackage{mwe}
\usepackage{caption}
\usepackage{lmodern}
\usepackage{natbib}
\usepackage{url}
\usepackage{comment}
\usepackage[utf8]{inputenc}
\usepackage[svgnames]{xcolor}
\usepackage{amsfonts,amssymb}
\usepackage{times,rotating}
\usepackage{graphicx,cite,calc}
\usepackage{mathtools}
\usepackage{amsmath,algorithm,algorithmic}
\usepackage{pgfplots}
\usepackage{abstract}
\usepackage{tikz,tikz-3dplot}
\usetikzlibrary{patterns}
\usetikzlibrary{fit,calc,positioning,decorations.pathreplacing,matrix,decorations.markings,shapes,spy}
\usetikzlibrary{arrows, positioning, calc,matrix,shadows,fadings,shapes.arrows,patterns,snakes}
\tikzstyle{line}=[draw]
\tikzstyle{arrow}=[draw, -latex] 
\usetikzlibrary{3d}
\usepackage{graphicx}
\usepackage{caption}
\usepackage{subcaption}

\pgfkeys{/tikz/.cd,
  arrow color/.initial=black!80!white,
  arrow color/.get=\arrowcol,
  arrow color/.store in=\arrowcol,
  items distance/.initial=2.5cm,
  items distance/.get=\itemdistance,
  items distance/.store in=\itemdistance,
  border color/.initial=black!80!white,
  border color/.get=\bordercol,
  border color/.store in=\bordercol,
  fill color/.initial=green!40!black!20,
  fill color/.get=\fillcol,
  fill color/.store in=\fillcol,
  brace color/.initial=black,
  brace color/.get=\bracecol,
  brace color/.store in=\bracecol,
  brace distance/.initial=5pt,
  brace distance/.get=\bracedistance,
  brace distance/.store in=\bracedistance,
}

\tikzset{my arrow/.style={
    single arrow, draw, minimum height=1.75cm,
    minimum width=2.5cm,
    single arrow head extend=0.1cm
  },
brace/.style={
    decoration={brace,mirror,raise=\bracedistance,amplitude=0.75em},
    decorate,
    draw=\bracecol,
    very thick,    
  },
brace_nonmirror/.style={
    decoration={brace,raise=\bracedistance,amplitude=0.75em},
    decorate,
    draw=\bracecol,
    very thick,    
  }
}

\tikzset{xcenter around/.style 2 args={execute at end picture={%
  \useasboundingbox let \p0 = (current bounding box.south west), \p1 = (current bounding box.north east),
                        \p2 = (#1), \p3 = (#2)
                    in
        ({min(\x2 + \x3 - \x1,\x0)},\y0) rectangle ({max(\x3 + \x2 - \x0,\x1)},\y1);
}}}

\newcommand{\beq}{\begin{equation}}
\newcommand{\eeq}{\end{equation}}
\newcommand{\beqa}{\begin{eqnarray}}
\newcommand{\eeqa}{\end{eqnarray}}
\newcommand{\bit}{\begin{itemize}}
\newcommand{\eit}{\end{itemize}}
\newcommand{\ben}{\begin{enumerate}}
\newcommand{\een}{\end{enumerate}}
\newcommand{\mc}{\mathcal}
\newcommand{\mb}{\mathbb}
\newcommand{\bed}{\begin{displaymath}}
\newcommand{\eed}{\end{displaymath}}

\newtheorem{theorem}{Theorem}

\newtheorem{lemma}[theorem]{Lemma}
\newtheorem{rem}[theorem]{Remark}

\begin{document}

\title{Convolutional Neural Associative Memories: Massive Capacity with Noise Tolerance}

\author{Amin Karbasi\\
\begin{tabular}{c}
E-mail: amin.karbasi@ethz.ch\\
Computer Science Department\\
Swiss Federal Institute of Technology Zurich, 8092 Zurich, Switzerland
\\
\\
Amir Hesam Salavati\\
E-mail: hesam.salavati@epfl.ch\\
Computer and Communication Sciences Department\\
Ecole Polytechnique Federale de Lausanne, Lausanne, 1015, Switzerland
\\
\\
Amin Shokrollahi\\
E-mail: amin.shokrollahi@epfl.ch\\    
\\
Computer and Communication Sciences Department\\
Ecole Polytechnique Federale de Lausanne, Lausanne, 1015, Switzerland\\
\end{tabular}}

\maketitle
\thispagestyle{empty}
\begin{abstract}
The task of a neural associative memory is to retrieve a set of previously memorized patterns from their noisy versions using a network of neurons. An ideal network should have the ability to 1) learn a set of patterns as they arrive, 2) retrieve the correct patterns from noisy queries, and 3) maximize the pattern retrieval capacity while maintaining the reliability in responding to queries.
The majority of work on neural associative memories has focused on designing networks capable of memorizing any set of randomly chosen patterns at the expense of limiting the retrieval capacity. 

In this paper, we show that if we target memorizing only those patterns that have inherent redundancy (i.e., belong to a subspace), we can obtain all the aforementioned properties. This is in sharp contrast with the previous work that could only improve one or two aspects at the expense of the third. More specifically, we propose framework based on a convolutional neural network along with an iterative algorithm that learns the redundancy among the patterns. The resulting network has a retrieval capacity that is \textit{exponential} in the size of the network. Moreover, the asymptotic error correction performance of our network is \textit{linear} in the size of the patterns. 
We then extend our approach to deal with patterns lie \textit{approximately} in a subspace. This extension allows us to memorize datasets containing natural patterns (e.g., images). Finally, we report experimental results on both synthetic and real datasets to support our claims.

\end{abstract}

\section{Introduction}
The ability of neuronal networks to memorize a large set of patterns and reliably retrieve them in the presence of noise, has attracted a large body of research over the past three decades to design artificial neural associative memories with similar capabilities. Ideally, a perfect neural associative memory should be able to \emph{learn} patterns, have a large pattern retrieval \emph{capacity} and be \emph{noise-tolerant}. This problem, called ''associative memory", is in spirit very similar to reliable information transmission faced in communication systems where the goal is to efficiently decode a set of transmitted patterns over a noisy channel. 
%

Despite this similarity and common methods deployed in both fields (e.g., graphical models, iterative algorithms, to name a few), we have witnessed a huge gap between the efficiency achieved by them. More specifically, by deploying modern coding techniques, it was shown that the number of reliably transmitted patterns over a noisy channel can be made \textit{exponential} in $n$, the length of the patterns. This was particularly achieved by imposing redundancy among transmitted patterns. In contrast, the maximum number of patterns that can be reliably memorized by most current neural networks scales \textit{linearly} in the size of the patterns. This is due to the common assumption that a neural network should be able to memorize \emph{any} subset of patterns drawn randomly from the set of all possible vectors of length $n$ (see, for example \citealp{hopfield}, \citealp{venkatesh}, \citealp{Jankowski}, \citealp{Muezzinoglu}).

Recently, \citet{KSS} suggested a new formulation of the problem where only a suitable set of patterns was considered for storing. To enforce the set of constraints, they formed a bipartite graph (as opposed to a complete graph considered in the earlier work) where one layer feeds the patterns to the network and the other takes into account the inherent structure. The role of bipartite graph is indeed similar to the Tanner graphs used in modern coding techniques \citep{tanner}. Using this model, \citet{KSS} provided evidence that the resulting network can memorize an exponential number of patterns at the expense of correcting only a \emph{single} error during the recall phase. By introducing a multi-layer structure, \citet{SK_ISIT2012} could further improve the error correction performance to \emph{constant} number of errors.
%



In this paper, similar to the model considered by \citet{KSS}, we only consider a set of patterns with weak minor components, i.e., patterns that lie in a subspace. By making use of this inherent redundancy
\begin{itemize}
\item We introduce the first convolutional neural associative network with provably exponential storage capacity.
\item We prove that our architecture can correct a linear fraction of errors. 
\item We develop an \textit{online} learning algorithm with the ability to learn patterns as they arrive. This property is specifically useful when the size of the dataset is massive and patterns can only be learned in a streaming manner. 
\item We extend our results to the case where patterns lie \textit{approximately} in a subspace. This extension in particular allows us to efficiently memorize datasets containing natural patterns.
\item We evaluate the performance of our proposed architecture and the learning algorithm through numerical simulations.
\end{itemize}
We provide rigorous analysis to support our claims. The storage capacity and error correction performance of our method is information-theoretically order optimum, i.e., no other method can significantly improve the results (except for constants). Our learning algorithm is an extension of the subspace learning method proposed by \citet{oja}, with an additional property of imposing the learned vectors to be \emph{sparse}. The sparsity is essential during the noise-elimination phase.

The remainder of this paper is organized as follows. In Section~\ref{sec:convolutional_related} we provide an overview of the related work in this area. In Section \ref{sec:convolutional_problem_folrmulation} we introduce our notation and formally state the problems that is the focus of this work, namely, learning phase, recall phase, and storage capacity. We present our learning algorithm in Section~\ref{section_learning} and our error correction method in Section \ref{sec:convolutional_recall}. Section~\ref{section_capacity} is devoted to the pattern retrieval capacity. We then report our experimental results on synthetic and natural datasets in Section~\ref{sec:experiments}. Finally, all the proofs are provided in Section~\ref{sec:analysis}.
\section{Related Work}\label{sec:convolutional_related}
The famous Hopfield network was among the first auto-associative neural mechanisms capable of learning a set of patterns and recalling them subsequently \citep{hopfield}. By employing the Hebbian learning rule \citep{hebb}, Hopfield considered a neural network of size $n$ with binary state neurons. It was shown by \citet{mceliece} that the capacity of a Hopfield network is bounded by $C=(n/2\log(n))$. Due to the low capacity of Hopfield networks, extension of associative memories to non-binary neural models has also been explored, with the hope of increasing the pattern retrieval capacity. In particular, \citet{Jankowski} investigated a complex-valued neural associative memory where each neuron can be assigned a multivalued state from the set of complex numbers. It was shown by \citet{Muezzinoglu} that the capacity of such networks can be increased to $C=n$ at the cost of a prohibitive weight computation mechanism. To overcome this drawback, a Modified Gradient Descent learning Rule (MGDR) was devised by \citet{Lee}. 

Recently, in order to increase the capacity and robustness, a line of work considered exploiting the inherent structure of the patterns. This is done by either making use of the correlations among the patterns or memorizing only those patterns that have some sort of redundancy. Note that they differ from the previous work in one important aspect: not any possible set of patterns is considered for learning, but only those with common structures. By employing neural cliques, \citet{gripon_sparse} were among the first to demonstrate that considerable improvements in the pattern retrieval capacity of Hopfield networks is possible, albeit still not passing the polynomial boundary on the capacity, i.e., $C=O(n^2)$. 

Similar idea was proposed by \citet{venkatesh_exponential} for learning semi-random patterns. This boost to the capacity is achieved by dividing the neural network into smaller fully interconnected \emph{disjoint} blocks. Using this idea, the capacity is increased to $\Theta \left(b^{n/b} \right)$, where $b = \omega( \ln n)$ is the size of clusters, Nonetheless, it was observed that this improvement comes at the price of limited noise tolerance capabilities.

By deploying higher order neural models, in contrast to the pairwise correlation considered in Hopfield networks, \citet{peretto} showed that the storage capacity can be improved to $C=O(n^{p-2})$, where $p$ is the degree of correlation. In such models, the state of the neurons not only depends on the state of their neighbors, but also on the correlations among them. However, the main drawback of this work lies in the prohibitive computational complexity of the learning phase. Recently, \citet{KSS} introduced a new model based on bipartite graphs to capture higher order (linear) correlations without the prohibitive computational complexity in the learning phase. The proposed model was further improved later \citep{kss_journal}, Under the assumption that the bipartite graph is fully known, sparse, and expander, the proposed algorithm by \citet{KSS} increased the pattern retrieval capacity to $C=O(a^n)$, for some $a > 1$. In addition to those restrictive assumptions, the performance of the recall phase was still below par.


In this paper, we introduce a convolutional neural network, capable of memorizing an exponential number of structured patterns while being able to correct a linear fraction of noisy neurons. Similar to the model considered by \citet{KSS}, we assume that patterns lie in a low dimensional subspace. Note that a network of size $n$, where each neuron can hold a finite number of states, is capable of memorizing at most an exponential number of patterns in $n$. Also, correcting a linear fraction of noisy nodes of the network is the best we can hope for. In addition, and more importantly in practice, we extend our results to the set of patterns that only approximately belong to a subspace.

%



It is worth mentioning that learning a set of input patterns with robustness against noise is not just the focus of neural associative memories. For instance, \citet{vincent} proposed an interesting approach to extract robust features in autoencoders. Their approach is based on \emph{artificially introducing} noise during the learning phase and let the network learn the mapping between the corrupted input and the correct version. This way, they shifted the burden from the recall phase to the learning phase. We, in contrast, consider another form of redundancy and enforce a suitable pattern structure that helps us design faster algorithms and derive necessary conditions that help us \textit{guarantee} to correct a linear fraction of noise without previously being exposed to.


Although our neural architecture is not technically considered a Deep Belief Network (DBN), it shares some similarities. DBNs are typically used to extract/classify features by the means of several consecutive stages (e.g., pooling, rectification, etc). Having multiple stages help the network to learn more interesting and complex features. An important class of DBNs are convolutional DBNs. The input layer (also known as the receptive field) is divided into multiple overlapping patches and the network extracts features from each patch \citep{jarrett}. Since we divide the input patterns into a few overlapping smaller clusters, our model is similar to those of convolutional DBNs. Furthermore, we also learn multiple \emph{features} (in our case dual vectors) from each patch where the feature extractions differ over different patches. This is indeed very similar to the approach proposed by \citet{le}. In contrast to convolutional DBNs, the focus of this work is not classification but rather recognition of the \emph{exact} patterns from their noisy versions. Moreover, in most DBNs, we not only have to find the proper dictionary for classification, but we also need to calculate the features for each input pattern. This alone increases the complexity of the whole system, especially if denoising is part of the objective. In our model, however, the dictionary is defined in terms of dual vectors. Consequently, previously memorized patterns are computationally easy to recognize as they yield the all-zero vector in the output of the feature extraction stage. In other words, a non-zero output can only happen if the input pattern is noisy. Another advantage of our model over DBNs is a much faster learning phase. More precisely, by using a single layer with overlapping clusters in our model the information diffuses gradually in the network. The same criteria is achieved in DBNs by constructing several stages \citep{socher}. 

\section{Problem Formulation}\label{sec:convolutional_problem_folrmulation}
In this section, we set our notation and formally define the learning phase, recall phase, and the storage capacity.
\subsection{Learning Phase} 
Throughout this paper, each pattern is denoted by an integer-valued vector $x = (x_1, x_2, \dots, x_n)$ of length $n$ where $x_i \in \mc{Q} = \{0,\dots,Q-1\}$ for $i=1,\dots,n$ and $Q$ is a non-negative integer. In words, the set $\mc{Q}$ could be thought of as the short term firing rate of neurons. Let $\{s_i\}_i^n$ denote the states of neurons in a neural network $G$. Each neuron updates its state based on the states of its neighbors. More precisely, a neuron $j$ first computes a weighted some $\sum_{i\in \mathcal{N}(j)} w_{j,i} s_i$ and then applies a nonlinear \textit{activation} function $f:\mathbb{R}\rightarrow \mc{Q}$, i.e., $$s_j = f\left(\sum_{i\in \mathcal{N}(j)} w_{j,i} s_i\right).$$ Here, $w_{j,i}$ is the weight of the neural connection between neurons $j$ and $i$, and $\mathcal{N}(j)$ denotes the neighbors of neuron $j$ in $G$. There are several possible  activation functions used in the literature including, but not limited to, linear, threshold, logistic, and tangent hyperbolic functions.

\begin{figure}
\centering
\begin{tikzpicture}[scale=0.9, transform shape]

\node at (-5.75,1.5)[rectangle,rounded corners,draw=black!80!blue!80,fill=black!40!blue!80,minimum size=8mm] (c1){\textcolor{white}{$y_1$}};
\node at (-4.25,1.5)[rectangle,rounded corners,draw=black!80!blue!80,fill=black!40!blue!80,minimum size=8mm] (c2){\textcolor{white}{$y_2$}};
\node at (-2.75,1.5)[rectangle,rounded corners,draw=black!80!blue!80,fill=black!40!blue!80,minimum size=8mm] (c3){\textcolor{white}{$y_3$}};

\node at (-6,-1.5)[circle,draw=black!40!orange!90,fill=black!10!orange!90,minimum size=11mm] (p1) {};
\node at (-4.3,-1.5)[circle,draw=black!40!orange!90,fill=black!10!orange!90,minimum size=11mm] (p2) {};
\node at (-2.6,-1.5)[circle,draw=black!40!orange!90,fill=black!10!orange!90,minimum size=11mm,font=\footnotesize] (p3) {$x_1^{(2)}$};

\node at (-4.25,2.75) (g1a) {\textcolor{black!40!blue!80}{$G^{(1)}$}};
\draw[brace_nonmirror,color=black!40!blue!80] (-6.25,2)--(-2.25,2);

\draw[dashed,draw=black!80!blue!80,rounded corners] (-6.75,-.75) rectangle (-1.75,-2.25);

\node at (-.65,1.5)[rectangle,rounded corners,draw=black!80!red!70,fill=black!30!red!90,minimum size=8mm] (c4) {\textcolor{white}{$y_4$}};
\node at (0.9,1.5)[rectangle,rounded corners,draw=black!80!red!70,fill=black!30!red!90,minimum size=8mm] (c5) {\textcolor{white}{$y_5$}};

\node at (-.9,-1.5)[circle,draw=black!40!orange!90,fill=black!10!orange!90,minimum size=11mm,font=\footnotesize] (p4) {$x_2^{(2)}$};
\node at (.8,-1.5)[circle,draw=black!40!orange!90,fill=black!10!orange!90,minimum size=11mm,font=\footnotesize] (p5) {$x_3^{(2)}$};

\draw[brace_nonmirror,color=black!30!red!90] (-1.15,2)--(1.4,2);
\node at (.125,2.75) (g2a) {\textcolor{black!30!red!90}{$G^{(2)}$}};

\draw[dashed,draw=black!80!red!70,rounded corners] (-3.3,-.5) rectangle (3.2,-2.5);

\node at (2.75,1.5)[rectangle,rounded corners,draw=black!80!green!80,fill=black!60!green!80,minimum size=8mm] (c6){\textcolor{white}{$y_6$}};
\node at (4.25,1.5)[rectangle,rounded corners,draw=black!80!green!80,fill=black!60!green!80,minimum size=8mm] (c7){\textcolor{white}{$y_7$}};
\node at (5.75,1.5)[rectangle,rounded corners,draw=black!80!green!80,fill=black!60!green!80,minimum size=8mm] (c8){\textcolor{white}{$y_8$}};

\node at (0,0)[yshift=-1.5cm,xshift=2.5cm,circle,draw=black!40!orange!90,fill=black!10!orange!90,minimum size=11mm,font=\footnotesize] (p6) {$x_4^{(2)}$};
\node at (0,0)[yshift=-1.5cm,xshift=4.2cm,circle,draw=black!40!orange!90,fill=black!10!orange!90,minimum size=11mm] (p7) {};
\node at (0,0)[yshift=-1.5cm,xshift=5.9cm,circle,draw=black!40!orange!90,fill=black!10!orange!90,minimum size=11mm] (p8) {};

\node at (4.25,2.75) (g3a) {\textcolor{black!60!green!80}{$G^{(3)}$}};
\draw[brace_nonmirror,color=black!60!green!80] (2.25,2)--(6.25,2);

\draw[dashed,draw=black!80!green!80,rounded corners] (1.6,-.75) rectangle (6.75,-2.25);

\draw[line,black!40!blue!80] (p1.north)--(c1.south);
\draw[line,black!40!blue!80] (p1.north)--(c2.south);
\draw[line,black!40!blue!80] (p2.north)--(c1.south);
\draw[line,black!40!blue!80] (p2.north)--(c3.south);
\draw[line,black!40!blue!80] (p3.north)--(c1.south);
\draw[line,black!40!blue!80] (p3.north)--(c3.south);
\draw[line,black!30!red!90] (p3.north)--(c4.south);
\draw[line,black!30!red!90] (p4.north)--(c4.south);
\draw[line,black!30!red!90] (p4.north)--(c5.south);
\draw[line,black!30!red!90] (p5.north)--(c4.south);
\draw[line,black!30!red!90] (p5.north)--(c5.south);
\draw[line,black!30!red!90] (p6.north)--(c5.south);
\draw[line,black!60!green!80] (p6.north)--(c6.south);
\draw[line,black!60!green!80] (p7.north)--(c6.south);
\draw[line,black!60!green!80] (p7.north)--(c7.south);
\draw[line,black!60!green!80] (p7.north)--(c8.south);
\draw[line,black!60!green!80] (p8.north)--(c7.south);
\draw[line,black!60!green!80] (p8.north)--(c8.south);

\node at (-8.5,1.5) (note1) {\textcolor{black!80!white!70}{Constraint neurons}};
\node at (-8.5,-1.5) (note2) {\textcolor{black!80!white!70}{Pattern neurons}};

\end{tikzpicture}
\caption{Bipartite graph $G$. In this figure we see three subpatterns $x^{(1)}, x^{(2)}, x^{(3)}$ along with corresponding clusters $G^{(1)},G^{(2)},G^{(2)}$. The subpattern $x^{(2)}$ has overlaps with both $G^{(1)}$ and $G^{(3)}$. The weights $w_{i,j}$ are chosen to ensure that $W\cdot x = 0$ for all patterns $x$ lying in a subspace. \label{fig:overlapping_clustered_network}}
\end{figure}
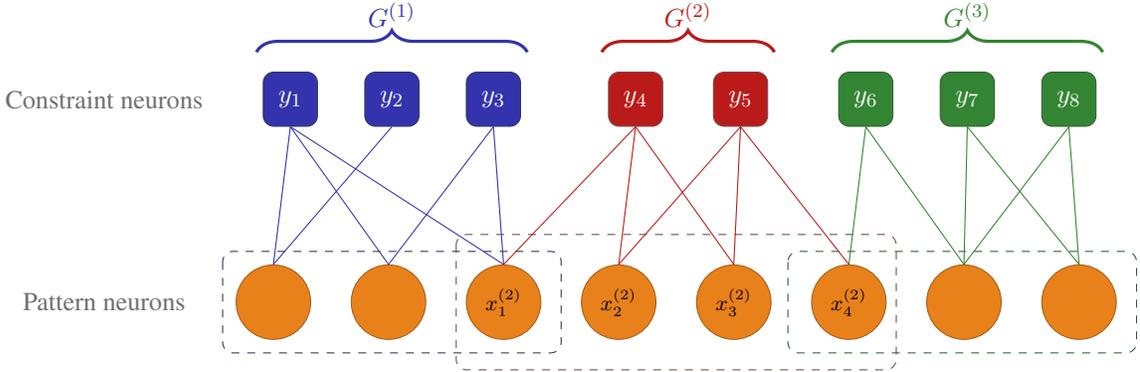

We denote the dataset of the patterns by the $C\times n$ dimensional matrix $\mc{X}$, where patterns are stored as the rows. Our goal in this work is to memorize patterns with strong \textit{local correlation} among the entries. More specifically, we divide the entries of each pattern $x$ into $L$ \emph{overlapping sub-patterns} of lengths $n_1,\dots,n_L$, so that $\sum n_i \geq n$. Note that due to overlaps, an entry in a pattern can be a member of multiple sub-patterns, as shown in Figure~\ref{fig:overlapping_clustered_network}. We denote the $i$-th sub-pattern by $$x^{(i)} = (x_1^{(i)}, x_2^{(i)}, \dots, x_{n_i}^{(i)}).$$ To enforce local correlations, we assume that the sub-patterns $x^{(i)}$ form a subspace of dimension $k_i<n_i$. This is done by imposing linear constraints on each cluster. 

These linear constraints are captured during the learning phase in the form of dual vectors. More specifically,
 we find a set of non-zero \textit{vectors} $w_1^{(i)}, w_2^{(i)}, \dots, w_{m_i}^{(i)}$ that are orthogonal to the set of sub-patterns $x^{(i)}$, i.e.,
\begin{equation}\label{eq:convolutional_orthogonal}
y_j^{(i)}=\langle w_j^{(i)} , x^{(i)}\rangle=0, \quad \forall j\in [m_i] \, \forall i\in [L],
\end{equation}
where $[q]$ denotes the set $\{1,2,\cdots, q\}$ and $\langle \cdot ,\cdot \rangle$ represents the inner product.

The weight matrix $W^{(i)}$ is constructed by placing all dual vectors next to each other, i.e., $$W^{(i)} = [w_1^{(i)}| w_2^{(i)}| \dots| w_{m_i}^{(i)}]^\top.$$ Equation~\eqref{eq:convolutional_orthogonal} can be written equivalently as 
\bed
W^{(i)} \cdot x^{(i)} = 0.
\eed
Cluster $i$ represents the bipartite graph $G^{(i)}$ with the connectivity matrix $W^{(i)}$. In the next section, we develop an iterative algorithm to learn the weight matrices $W^{(1)},\dots,W^{(L)}$, while encouraging sparsity within each connectivity matrix $W^{(i)}$.

One can easily map the local constraints imposed by the $W^{(i)}$'s into a global constraint by introducing a global weight matrix $W$ of size $m\times n$. The first $m_1$ rows of the matrix $W$ correspond to the constraints in the first cluster, rows $m_1+1$ to $m_1+m_2$ correspond to the constraints in the second cluster, and so forth. Hence, by inserting zero entries at proper positions, we can construct the global constraint matrix $W$. We will use both the local and global connectivity matrices to eliminate noise in the recall phase.

\subsection{Recall Phase} 
In the recall phase a noisy version, say $\hat{x}$, of an already learned pattern $x\in \mc{X}$ is given. Here, we assume that the noise is an additive vector of size $n$, denoted by $e$, whose entries assume values independently from $\{-1,0,+1\}$\footnote{In our experiments, we have considered larger integer values for noise as well, i.e., $\{-q,\dots,0,\dots,q\}$, for some $q \in \mb{N}$. The $\pm 1$ noise model here is considered to simplify the notations and analysis.} with corresponding probabilities $p_{-1} =p_{+1}=p_e/2$ and $p_0=1-p_e$. In other words, each entry of the noise vector is set to $\pm 1$ with probability $p_e$. The $\pm 1$ values are chosen to simplify the analysis. Our approach can be easily extended to other integer-valued noise models. 

We denote by $e^{(i)}$, the realization of noise on the sub-pattern $x^{(i)}$. In formula, $\hat{x}=x+e$.\footnote{Note that since entries of $\hat{x}$ should be between $0$ and $Q-1$, we cap values below $0$ and above $Q-1$ to $0$ and $Q-1$, respectively.} Note that $W\cdot \hat{x} = W\cdot e$ and $W^{(i)} \cdot \hat{x}^{(i)} = W^{(i)} \cdot e^{(i)}$. Therefore, the goal in the recall phase is to remove the noise $e$ and recover the desired pattern $x$. This task will be accomplished by exploiting the facts that a) we have chosen the set of patterns $\mc{X}$ to satisfy the set of constraints $W^{(i)} \cdot x^{(i)} = 0$ and b) we opted for sparse neural graphs $G^{(i)}$ during the learning phase. Based on these two properties, we develop the first recall algorithm that corrects a linear fraction of noisy entries. 

\subsection{Capacity} 
The last issue we look at in this work is the retrieval capacity $C$ of our proposed method. Retrieval or critical storage capacity is defined as the maximum number of patterns that a neural network is able to store without having (significant) errors in returned answers during the recall phase. Hence, the storage capacity $C(n)$ is usually measured in terms of the network size $n$. It is well known that the retrieval capacity is affected by certain considerations about the neural network, including the range of values or states for the patterns, inherent structure of patterns, and topology of neural networks. In this work, we show that a careful combination of patterns' structure and neural network topology leads to an exponential storage capacity in the size of the network. 
\section{The Learning Algorithm}\label{section_learning}
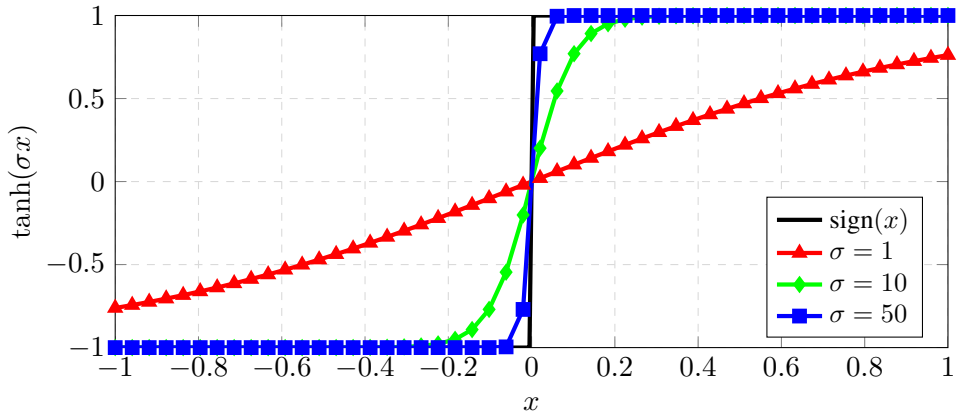
\begin{figure}[t]
\begin{center}
\begin{tikzpicture}[scale=1.0]
\begin{axis}[
	 scaled ticks=false, tick label style={/pgf/number format/fixed},
	 width=\textwidth, height=6cm,     
	 ylabel=$\tanh(\sigma x)$,
	grid = major,
        grid style={dashed, gray!30},
	 xlabel=$x$,
 	 xmin=-1,
	 xmax=1,
 	 ymin=-1,
 	 ymax=1,
	legend style={
		cells={anchor=west},
		legend pos=south east,
		font = \small,
	}	
    ]
    \addplot[domain=-1:1,samples=200, black, ultra thick] {x/abs(x)};
    \addplot[domain=-1:1,samples=50, red, ultra thick,mark=triangle*] {tanh(x)};
    \addplot[domain=-1:1,samples=50, green, ultra thick,mark=diamond*] {tanh(10*x)};
    \addplot[domain=-1:1,samples=50, blue, ultra thick,mark=square*] {tanh(50*x)};
    \legend{$\hbox{sign}(x)$\\ $\sigma = 1$\\ $\sigma = 10$\\ $\sigma = 50$\\};
\end{axis}
\end{tikzpicture}
\end{center}
\begin{center}
\caption{Approximation of $\hbox{sign}(x)$ by $\tanh(\sigma x)$. As we increase the value of $\sigma$ the approximation becomes more accurate. 
 \label{fig:tanh_approximation}}
\end{center}
\end{figure}

In this section, we develop an algorithm for learning the weight matrix $W^{(\ell)}$ of a given cluster $\ell$. 
By our assumptions, the sub-patterns lie in a subspace of dimension $k_\ell \leq n_\ell$. Hence we can adopt the iterative algorithm proposed by \citet{oja_proof} and \citet{xu} to learn the corresponding null space.
However, in order to ensure the success of the denoising algorithm proposed in Section~\ref{sec:convolutional_recall}, we require $W^{(l)}$ to be sparse. To this end, the objective function shown below has a penalty term to encourage sparsity. 
 Furthermore, we are not seeking an orthogonal \emph{basis} as in the approach proposed by \citet{xu}. Instead, we wish to find $m_\ell$ vectors $w^{(\ell)}$ that are orthogonal to the (sub-) patterns. Hence, the optimization problem for finding a constraint \emph{vector} $w^{(\ell)}$ can be formulated as follows:
%
%
\begin{eqnarray}\label{main_problem_modified}
&\min_{w^{(\ell)}}& \sum_{x \in \mc{X}} \vert \langle x^{(\ell)} , w^{(\ell)} \rangle \vert^2 + \eta g(w^{(\ell)}), \label{main_problem_modified}\\
 &\text{s.t. }& \Vert w^{(\ell)} \Vert_2 = 1.\label{eq:constraint}
\end{eqnarray}
In the above problem, $x^{(\ell)}$ is a sub-pattern of $x$ drawn from the training set $\mc{X}$, $\langle \cdot ,\cdot \rangle$ indicates the inner product, $\eta$ is a positive constant, and $g(\cdot)$ is the penalty term to favor sparse results. In this paper, we consider 
$$g(w^{(\ell)}) = \sum_{i = 1}^n \tanh(\sigma (w^{(\ell)}_i)^2).$$
It is easy to see that for large $\sigma$, the function $\tanh(\sigma (w^{(\ell)}_i)^2)$ approximates $|\hbox{sign}(w^{(\ell)}_i)|$ (as shown in Figure \ref{fig:tanh_approximation}). Therefore, the larger $\sigma$ gets, the closer $g(w^{(\ell)})$ will be to $\Vert \cdot \Vert_0$. Another popular choice, widely used in compressed sensing (see, for example, \citealp{donoho_sens} and \citealp{candes}), is to pick $g(w^{(\ell)})=\Vert w^{(\ell)} \Vert_1$. Note that the optimization problem \eqref{main_problem_modified} without the constraint \eqref{eq:constraint} has the trivial solution $w^{(\ell)} = \underline{0}$ where $\underline{0}$ is the all-zero vector.

 
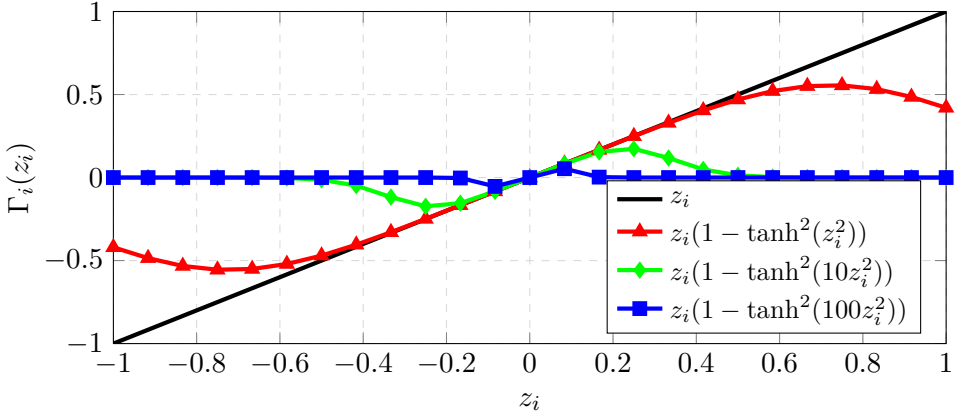
\begin{figure}[t]
\begin{center}
\begin{tikzpicture}[scale=1.0]
\begin{axis}[
	 scaled ticks=false, tick label style={/pgf/number format/fixed},
	 width=\textwidth, height=6cm,     
	 ylabel=$\Gamma_i(z_i)$,
	grid = major,
        grid style={dashed, gray!30},
	 xlabel=$z_i$,
 	 xmin=-1,
	 xmax=1,
 	 ymin=-1,
 	 ymax=1,
	legend style={
		cells={anchor=west},
		legend pos=south east,
		font = \small,
	}	
    ]
    \addplot[domain=-1:1, black, ultra thick] {x};
    \addplot[domain=-1:1, red, ultra thick,mark=triangle*] {x*(1-(tanh(x^2))^2)};
    \addplot[domain=-1:1, green, ultra thick,mark=diamond*] {x*(1-(tanh(10*x^2))^2)};
    \addplot[domain=-1:1, blue, ultra thick,mark=square*] {x*(1-(tanh(100*x^2))^2)};
    \legend{$z_i$\\ $z_i(1-\tanh^2(z_i^2))$\\ $z_i(1-\tanh^2(10z_i^2))$\\ $z_i(1-\tanh^2(100z_i^2))$\\};
\end{axis}
\end{tikzpicture}
\end{center}
\begin{center}
\caption{
The sparsity penalty $\Gamma_i(z_i)$ suppresses small values of $z_i$ towards zero. Note that as $\sigma$ gets larger, the support of $\Gamma_i(z_i)$ gets smaller.
 \label{sparsity_penalty}}
\end{center}
\end{figure}

To minimize the objective function shown in \eqref{main_problem_modified} subject to the norm constraint \eqref{eq:constraint} we use stochastic gradient descent and follow a similar approach to that of \citet{xu}. By calculating the derivative of the objective function and considering the updates required for each randomly picked pattern $x$, we will obtain the following iterative algorithm:
\begin{eqnarray}
y^{(\ell)}(t) &=& \langle x^{(\ell)}(t), w^{(\ell)}(t)\rangle, \label{SGA_cost_modified}\\
 \tilde{w}^{(\ell)}(t+1) &=& w^{(\ell)}(t) - \alpha_t \left(2 y^{(\ell)}(t) x^{(\ell)}(t) + \eta\Gamma(w^{(\ell)}(t))\right),\label{SGA_learning_rule_modified}\\
 w^{(\ell)}(t+1) &=& \frac{\tilde{w}^{(\ell)}(t+1)}{\Vert \tilde{w}^{(\ell)}(t+1) \Vert_2}.\label{w(t)}
\end{eqnarray}
In the above equations, $t$ is the iteration number, $x^{(\ell)}(t)$ is the subpattern of a pattern $x(t)$ drawn at iteration $t$, $\alpha_t$ is a small positive constant, and $\Gamma(w^{(\ell)}) = \nabla g(w^{(\ell)})$ is the gradient of the penalty term. The function $\Gamma(w^{(\ell)})$ encourages sparsity. To see why, consider the $i$-th entry of $\Gamma(\cdot)$, namely, 
\begin{eqnarray*}
\Gamma_i(z_i) &=& \frac{\partial g(z)}{\partial z_i}\\ &=& 2\sigma z_i (1-\tanh^2(\sigma z_i^2)).
\end{eqnarray*}
Note that $\Gamma_i\simeq 2 \sigma z_i$ for relatively small values of $z_i$, and $\Gamma_i \simeq 0$ for larger values of $z_i$ (see Figure \ref{sparsity_penalty}). 
%

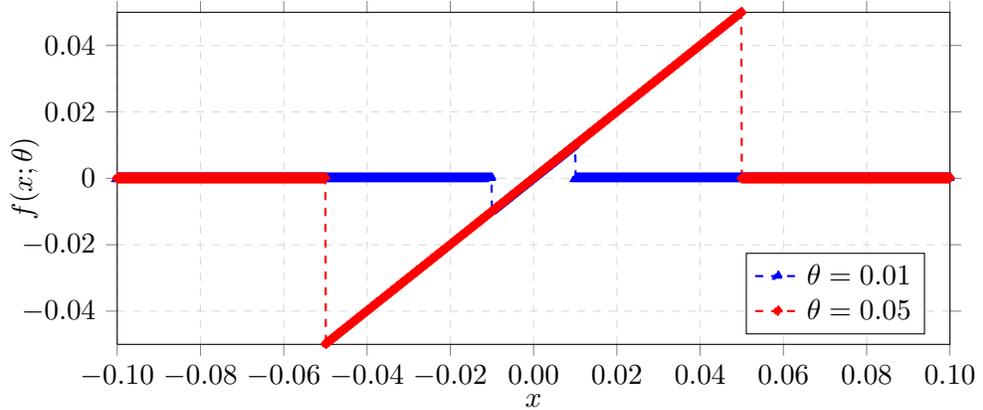
\begin{figure}[h]
\begin{center}
\centering
\begin{tikzpicture}
    \begin{axis}[
  	 scaled ticks=false, tick label style={/pgf/number format/fixed},
        width=\textwidth, height=6cm,     
        grid = major,
        grid style={dashed, gray!30},
        xmin=-0.1,     
        xmax=0.1,    
        ymin=-0.05,     
        ymax=0.05,   
        axis background/.style={fill=white},
        xlabel=$x$,
        ylabel=$f(x;\theta)$,
		x tick label style={
        /pgf/number format/.cd,
        fixed,
        fixed zerofill,
        precision=2,
        /tikz/.cd
    	 },
        tick align=outside,
	legend style={
		cells={anchor=west},
		legend pos=south east,
		},
	]
\addplot [color=blue,thick,dashed,mark=triangle*] table{soft_threshold_theta_001.dat};
\addplot [color=red,thick,dashed,mark=diamond*] table{soft_threshold_theta_005.dat};
\legend{$\theta = 0.01$\\$\theta = 0.05$\\}
\end{axis} 

\end{tikzpicture}
\end{center}
\begin{center}
\caption{The soft threshold function $f(x,\theta)$ for two different values of $\theta$.
 \label{fig:soft_threshold_func}}
\end{center}
\end{figure}

Thus, for proper choices of $\eta$ and $\sigma$, equation~(\ref{SGA_learning_rule_modified}) suppresses small entries of $w^{(\ell)}(t)$ towards zero and favors sparser results. To further simplify the iterative equations \eqref{SGA_cost_modified}, \eqref{SGA_learning_rule_modified}, \eqref{w(t)} we approximate the function $\Gamma(w^{(\ell)}(t))$ with the following threshold function (shown in Figure \ref{fig:soft_threshold_func}):

\bed
\Gamma_i(z_i,\theta_t) = \left\{ \begin{array}{ll}
   z_i & \mbox{if $\vert z_i \vert \leq \theta_t$};\\    
    0 & \mbox{otherwise}.\end{array} \right.
\eed
%
where $\theta_t$ is a small positive threshold. 
Following the same approach taken by \citet{oja_proof} we assume that $\alpha_t$ is small enough so that equation~(\ref{w(t)}) can be expanded as powers of $\alpha_t$. Also note that the inner product $\langle w^{(\ell)}(t), \Gamma(w^{(\ell)}(t),\theta_t)\rangle$ is small so in the power expansion we can omit the term $\alpha_t \eta \left(\langle w^{(\ell)}(t), \Gamma(w^{(\ell)}(t),\theta_t)\rangle\right) w^{(\ell)}(t).$

 By applying the above approximations we obtain an iterative learning algorithm shown in Algorithm~\ref{algo_learning}. In words, $y^{(\ell)}(t)$ is the projection of $x^{(\ell)}(t)$ onto $w^{(\ell)}(t)$. If for a given data vector $x^{(\ell)}(t)$ the projection $y^{(\ell)}(t)$ is non-zero, then the weight vector will be updated in order to reduce this projection.
\begin{algorithm}[b]
\caption{Iterative Learning}
\label{algo_learning}
\begin{algorithmic}[1]
\REQUIRE{Dataset $\mc{X}$ with $|\mc{X}|=C$, stopping point $\varepsilon$.}
\ENSURE{$w^{(\ell)}$}
\WHILE{ $\frac{1}{C} \sum_{x \in \mc{X}} \vert \langle x^{(\ell)}(t) , w^{(\ell)}(t) \rangle \vert^2 >\varepsilon$ }
\STATE Choose pattern $x(t)$ uniformly at from $\mc{X}$.
\STATE Compute $y^{(\ell)}(t) = \langle x^{(\ell)}(t) , w^{(\ell)}(t) \rangle $.
\STATE Update $w^{(\ell)}(t)$ as follows 
$$w^{(\ell)}(t)= w^{(\ell)}(t-1) - \alpha_t \left(y^{(\ell)}(t) \left(x^{(\ell)}(t) - \frac{y^{(\ell)}(t) w^{(\ell)}(t-1)}{\Vert w^{(\ell)}(t-1)\Vert_2^2}\right) + \eta\Gamma(w^{(\ell)}(t-1),\theta_t)\right)$$
\STATE $t \leftarrow t+1$.
\ENDWHILE
\end{algorithmic}
\end{algorithm}
%

\subsection{Convergence Analysis}\label{sec:convergence_learning}
Our main idea for proving the convergence of the learning algorithm is to consider the \textit{learning cost function} defined as follows:
\bed
E(t) = E(w^{(\ell)}(t)) =\frac{1}{C} \sum_{\mu =1}^{C} \left( \langle w^{(\ell)}(t), x^\mu \rangle \right)^2
\eed
We show that as we gradually learn patterns from the data set $\mc{X}$, the cost function $E(t)$ goes to zero. In order to establish this result we need to specify the learning rate $\alpha_t$. In what follows, we assume that $\alpha_t = \Omega(1/t)$ so that $\sum_t \alpha_t \rightarrow \infty$ and $\sum_t \alpha_t^2 < \infty$. 
%
%
%
We first show that the weight vector $w^{(\ell)}(t)$ never becomes zero, i.e., $\Vert w^{(\ell)}(t) \Vert_2 > 0$ for all $t$.
\begin{lemma}\label{lem:avoiding_all_zero}
Assume that $\Vert w^{(\ell)}(0) \Vert_2 > 0$ and $\alpha_0  < 1/\eta$. Then for all iterations $t$, we have $\alpha_t < \alpha_0  < 1/\eta$ and $\Vert w^{(\ell)}(t) \Vert_2 > 0$.
%
\end{lemma}
As mentioned earlier, all the proofs are given in Section \ref{sec:analysis}. 

The above lemma ensures that if we reach $E(t)=0$ for some iteration $t$, it is not the case that $w^{(\ell)}(t) = \underline{0}$.
Next, we prove the convergence of Algorithm \ref{algo_learning} to a minimum $\hat{w}^{(\ell)}$ for which $E(\hat{w}^{(\ell)}) = 0$. 
\begin{theorem}\label{lemma_convergence}
Under the conditions of Lemma~\ref{lemma_convergence}, the learning Algorithm \ref{algo_learning} converges to a local minimum $\hat{w}^{(\ell)}$ for which $E(\hat{w}^{(\ell)}) = 0$. Moreover $\hat{w}^{(\ell)}$ is orthogonal to all the patterns in the data set $\mc{X}$.
%
%
\end{theorem}
We should note here that a similar convergence result can be proven without introducing the penalty term $g(w^{(\ell)})$. However, our recall algorithm crucially depends on the sparsity level of learned $w^{(\ell)}$'s. As a consequence we encouraged sparsity by adding the penalty term $g(w^{(\ell)})$. Our experimental results in Section \ref{sec:experiments} show that in fact this strategy works perfectly and the learning algorithm results in sparse solutions.

In order to find $m_\ell$ constraints required by the learning phase, we need to run Algorithm \ref{algo_learning} \emph{at least} $L$ times. In practice, we can perform this process in parallel, to speed up the learning phase. It is also more meaningful from a biological point of view, as each constraint neuron can act independently from the others. Although running Algorithm~\ref{algo_learning} in parallel may result in redundant constraints, our experimental results show that by starting from different random initial points, the algorithm converges to linearly independent constraints almost surely.

\section{Recall Phase}\label{sec:convolutional_recall}
Once the learning phase is finished, the weights of the neural graphs are fixed. Thus, during the recall phase we assume that the connectivity matrix for each cluster $i$ (denoted by $W^{(i)}$) has been learned and satisfies \eqref{eq:convolutional_orthogonal}.

The recall phase of our proposed model consists of two parts: intra-cluster and inter-cluster. During the intra-cluster part, clusters try to remove noise from their own sub-patterns. As we will see shortly, each cluster succeeds in correcting a single error with high probability. Such individual error correction performance is fairly limited. The inter-cluster part capitalizes on the overlap among clusters to improve the overall performance of the recall phase. In what follows, we describe both parts in more details.


\subsection{Intra-cluster Recall Algorithm}\label{section_recall_intra_cluster} 
For the intra-cluster part, shown in Algorithm \ref{algo:correction}, we exploit the fact that the connectivity matrix of the
neural network in each cluster is sparse and orthogonal to the memorized patterns. As a result we have $W^{(\ell)} (x^{(\ell)} + e^{(\ell)}) = W^{(\ell)} e^{(\ell)}$ where $e^{(\ell)}$ is the noise added to the sub-pattern $x^{(\ell)}$. 


Algorithm \ref{algo:correction} performs a series of \emph{forward} and \emph{backward} iterations to remove $e^{(\ell)}$. At each iteration, the pattern neurons decide locally whether to update their current state or not: if the amount of feedback received by a pattern neuron exceeds a threshold, the neuron updates its state, and remains intact, otherwise.\footnote{In order to maintain the current value of a neuron, we can add self-loops to pattern neurons in Figure \ref{fig:overlapping_clustered_network}. the self-loops are not shown in the figure for the sake of clarity).}

\begin{algorithm}[t]
\caption{Intra-cluster Error Correction} 
\label{algo:correction}
\begin{algorithmic}[1]
\REQUIRE{Training set $\mc{X}$, threshold $\varphi$, iteration $t_{\max}$}
\ENSURE{$x^{(\ell)}_1,x^{(\ell)}_2,\dots,x^{(\ell)}_{n_\ell}$}
\FOR{$t = 1 \to t_{\max}$} 
\STATE \emph{Forward iteration:} Calculate the weighted input sum $ h_i^{(\ell)} = \sum_{j=1}^{n_\ell} W^{(\ell)}_{ij} x^{(\ell)}_j,$ for each neuron $y^{(\ell)}_i$ and set
$y^{(\ell)}_i = \hbox{sign}(h^{(\ell)}_i).$\footnotemark
\STATE \textit{Backward iteration:} Each neuron $x^{(\ell)}_j$ computes $$g^{(\ell)}_j = \frac{\sum_{i=1}^{m_\ell} W^{(\ell)}_{ij} y^{(\ell)}_i}{\sum_{i = 1}^{m_\ell}|W^{(\ell)}_{ij}|}.$$
\STATE Update the state of each pattern neuron $j$ according to $$x^{(\ell)}_j = x^{(\ell)}_j - \hbox{sign}(g^{(\ell)}_j)$$
only if $|g^{(\ell)}_j| > \varphi$.
 \STATE $t \gets t + 1$
\ENDFOR
\end{algorithmic}
\end{algorithm}
\footnotetext{In practice, we usually set $y^{(\ell)}_i = \hbox{sign}(h^{(\ell)}_i)$ only if $\vert h^{(\ell)}_i \vert > \psi$, where $\psi$ is a small positive threshold.}

In order to state our results, we need to define the \textit{degree distribution polynomial} (from the node perspective). More precisely, let $\Lambda^{(\ell)}_i$ be the fraction of \emph{pattern neurons} with degree $i$ in cluster $G^{(\ell)}$ and define $\Lambda^{(\ell)}(x) = \sum_i \Lambda^{(\ell)}_i x^i$ to be the degree distribution polynomial for the pattern neurons in cluster $\ell$. In principle $\Lambda^{(\ell)}(x)$ encapsulates all the information we need to know regarding cluster $G^{(\ell)}$, namely, the degree distribution. The following theorem provides a lower bound on the average probability of correcting a single erroneous pattern neuron by each cluster.

\begin{theorem}\label{theorem_algo_recall_within}
If we sample the neural graph randomly from $\Lambda^{(\ell)}(x)$, and let $\varphi \rightarrow 1$, then Algorithm~\ref{algo:correction} can correct (at least) a single error in cluster $G^{(\ell)}$ with probability at least
\bed
P_c^{(\ell)} = \left(1- \Lambda^{(\ell)}\left(\frac{\bar{d}_{\ell}}{m_\ell} \right) \right)^{n_\ell-1},
\eed
where $\bar{d}_{\ell}$, $n_\ell$, and $m_\ell$ are the average degree of pattern neurons, the number of pattern neurons, and the number of constraint neurons in cluster $G^{(\ell)}$, respectively.
%
\end{theorem}
To gain some intuition we can further simplify the expression in the above theorem as follows
\beqa
P_c^{(\ell)} &\geq& 
\left(1- \left(\frac{\bar{d}_{\ell}}{m_\ell} \right)^{d^{(\ell)}_{\min}} \right)^{n_\ell-1}\nonumber,
\eeqa
where $d^{(\ell)}_{\min}$ is the minimum degree of pattern neurons in cluster $\ell$ and we assumed that $d^{(\ell)}_{\min}\geq 1$. This shows the significance of having high-degree pattern neurons. In the extreme case of $d^{(\ell)}_{\min} = 0$
%
%
then we obtain the trivial bound of $P^{(\ell)}_c \geq (1-\Lambda^{(\ell)}_0)^{n_\ell-1}$, where $\Lambda^{(\ell)}_0$ is the fraction of pattern neurons with degree equal to $0$. In particular, for large $n_\ell$ we obtain $P^{(\ell)}_c \geq e^{-D^{(\ell)}_0}$, where $D^{(\ell)}_0 = \Lambda^{(\ell)}_0 n_\ell$ is the total number of pattern neurons with degree $0$. Thus, even if only a single pattern neuron has a zero degree, probability of correcting a single error drops significantly.
%

While Theorem \ref{theorem_algo_recall_within} provides a lower bound on the probability of correcting a single error when the connectivity graph is sampled according to the degree distribution polynomial $\Lambda^{(\ell)}(x)$, the following lemma shows that under mild conditions (that depends on the neighborhood relationship among neurons), Algorithm~\ref{algo:correction} will correct a single input error with probability $1$.
%

\begin{lemma}\label{corrolary_single_error_no_two_column}
If no two pattern neurons share the exact same neighborhood in cluster $G^{(\ell)}$, and as $\varphi \rightarrow 1$, Algorithm~\ref{algo:correction} corrects (at least) a single error.
\end{lemma}
%

For the remaining of the paper we let $P_c = \mb{E}_\ell \left(P_c^{(\ell)} \right)$ denote this average probability of correcting one error averaged over all clusters. Lemma~\ref{corrolary_single_error_no_two_column} suggests that nuder mild conditions, and in fact in many practical settings discussed later, $P_c$ is close to $1$. Thus, from now on we pessimistically assume that if there is a single error in a given cluster, Algorithm~\ref{algo:correction} corrects it with probability $P_c$ and declares a failure if there are more than one error.

\subsection{Inter-cluster Recall Algorithms}
\begin{figure}
\begin{subfigure}[b]{.5\textwidth}
\begin{tikzpicture}[scale=.5, transform shape]

\node at (-5.75,1.5)[rectangle,rounded corners,draw=black!80!white!80,fill=black!90!white!60,minimum size=8mm] (c1){\textcolor{white}{}};
\node at (-4.25,1.5)[rectangle,rounded corners,draw=black!80!white!80,fill=black!90!white!60,minimum size=8mm] (c2){\textcolor{white}{}};
\node at (-2.75,1.5)[rectangle,rounded corners,draw=black!80!white!80,fill=black!90!white!60,minimum size=8mm] (c3){\textcolor{white}{}};

\node at (-6,-1.5)[circle,draw=black!40!orange!90,fill=black!10!orange!90,minimum size=11mm] (p1) {};
\node at (-4.3,-1.5)[circle,draw=black!40!red!90,fill=black!10!red!90,minimum size=11mm] (p2) {};
\node at (-2.6,-1.5)[circle,draw=black!40!red!90,fill=black!10!red!90,minimum size=11mm,font=\footnotesize] (p3) {$$}; 

\node[font=\LARGE] at (-4.25,3) (g1a) {\textcolor{black!80!white!80}{$G^{(1)}$}};
\draw[brace_nonmirror,color=black!80!white!80] (-6.25,1.5)--(-2.25,1.5);

\draw[dashed,black!90!white!60,rounded corners] (-6.75,-.75) rectangle (-1.75,-2.25);

\node at (-.65,1.5)[rectangle,rounded corners,draw=black!80!white!80,fill=black!90!white!60,minimum size=8mm] (c4) {\textcolor{white}{}};
\node at (0.9,1.5)[rectangle,rounded corners,draw=black!80!white!80,fill=black!90!white!60,minimum size=8mm] (c5) {\textcolor{white}{}};

\node at (-.9,-1.5)[circle,draw=black!40!orange!90,fill=black!10!orange!90,minimum size=11mm,font=\footnotesize] (p4) {};
\node at (.8,-1.5)[circle,draw=black!40!orange!90,fill=black!10!orange!90,minimum size=11mm,font=\footnotesize] (p5) {};

\draw[brace_nonmirror,color=black!80!white!80] (-1.15,1.5)--(1.4,1.5);
\node [font=\LARGE] at (.125,3) (g2a) {\textcolor{black!80!white!80}{$G^{(2)}$}};

\draw[dashed,black!90!white!60,rounded corners] (-3.3,-.5) rectangle (3.2,-2.5);

\node at (2.75,1.5)[rectangle,rounded corners,draw=black!80!white!80,fill=black!90!white!60,minimum size=8mm] (c6){\textcolor{white}{}};
\node at (4.25,1.5)[rectangle,rounded corners,draw=black!80!white!80,fill=black!90!white!60,minimum size=8mm] (c7){\textcolor{white}{}};
\node at (5.75,1.5)[rectangle,rounded corners,draw=black!80!white!80,fill=black!90!white!60,minimum size=8mm] (c8){\textcolor{white}{}};

\node at (0,0)[yshift=-1.5cm,xshift=2.5cm,circle,draw=black!40!red!90,fill=black!10!red!90,minimum size=11mm,font=\footnotesize] (p6) {};
\node at (0,0)[yshift=-1.5cm,xshift=4.2cm,circle,draw=black!40!orange!90,fill=black!10!orange!90,minimum size=11mm] (p7) {};
\node at (0,0)[yshift=-1.5cm,xshift=5.9cm,circle,draw=black!40!orange!90,fill=black!10!orange!90,minimum size=11mm] (p8) {};

\node[font=\LARGE] at (4.25,3) (g3a) {\textcolor{black!80!white!80}{$G^{(3)}$}};
\draw[brace_nonmirror,color=black!80!white!80] (2.25,1.5)--(6.25,1.5);

\draw[dashed,black!90!white!60,rounded corners] (1.6,-.75) rectangle (6.75,-2.25);

\draw[line,black!90!white!60] (p1.north)--(c1.south);
\draw[line,black!90!white!60] (p1.north)--(c2.south);
\draw[line,black!90!white!60] (p2.north)--(c1.south);
\draw[line,black!90!white!60] (p2.north)--(c3.south);
\draw[line,black!90!white!60] (p3.north)--(c1.south);
\draw[line,black!90!white!60] (p3.north)--(c3.south);
\draw[black!90!white!60] (p3.north)--(c4.south);
\draw[black!90!white!60] (p4.north)--(c4.south);
\draw[black!90!white!60] (p4.north)--(c5.south);
\draw[black!90!white!60] (p5.north)--(c4.south);
\draw[black!90!white!60] (p5.north)--(c5.south);
\draw[black!90!white!60] (p6.north)--(c5.south);
\draw[line,black!60!blue!80] (p6.north)--(c6.south);
\draw[line,black!60!blue!80] (p7.north)--(c6.south);
\draw[line,black!60!blue!80] (p7.north)--(c7.south);
\draw[line,black!60!blue!80] (p7.north)--(c8.south);
\draw[line,black!60!blue!80] (p8.north)--(c7.south);
\draw[line,black!60!blue!80] (p8.north)--(c8.south);

\end{tikzpicture}
\caption{Initial step}
\end{subfigure}\quad
\begin{subfigure}[b]{.5\textwidth}
\begin{tikzpicture}[scale=.5, transform shape]

\node at (-5.75,1.5)[rectangle,rounded corners,draw=black!80!white!80,fill=black!90!white!60,minimum size=8mm] (c1){\textcolor{white}{}};
\node at (-4.25,1.5)[rectangle,rounded corners,draw=black!80!white!80,fill=black!90!white!60,minimum size=8mm] (c2){\textcolor{white}{}};
\node at (-2.75,1.5)[rectangle,rounded corners,draw=black!80!white!80,fill=black!90!white!60,minimum size=8mm] (c3){\textcolor{white}{}};

\node at (-6,-1.5)[circle,draw=black!40!orange!90,fill=black!10!orange!90,minimum size=11mm] (p1) {};
\node at (-4.3,-1.5)[circle,draw=black!40!red!90,fill=black!10!red!90,minimum size=11mm] (p2) {};
\node at (-2.6,-1.5)[circle,draw=black!40!red!90,fill=black!10!red!90,minimum size=11mm,font=\footnotesize] (p3) {$$};

\node[font=\LARGE] at (-4.25,3) (g1a) {\textcolor{black!80!white!80}{$G^{(1)}$}};
\draw[brace_nonmirror,color=black!80!white!80] (-6.25,1.5)--(-2.25,1.5);

\draw[dashed,black!80!white!80,rounded corners] (-6.75,3.5) rectangle (-1.75,-2.25);

\node at (-.65,1.5)[rectangle,rounded corners,draw=black!80!white!80,fill=black!90!white!60,minimum size=8mm] (c4) {\textcolor{white}{}};
\node at (0.9,1.5)[rectangle,rounded corners,draw=black!80!white!80,fill=black!90!white!60,minimum size=8mm] (c5) {\textcolor{white}{}};

\node at (-.9,-1.5)[circle,draw=black!40!orange!90,fill=black!10!orange!90,minimum size=11mm,font=\footnotesize] (p4) {};
\node at (.8,-1.5)[circle,draw=black!40!orange!90,fill=black!10!orange!90,minimum size=11mm,font=\footnotesize] (p5) {};

\draw[brace_nonmirror,color=black!80!white!80] (-1.15,1.5)--(1.4,1.5);
\node [font=\LARGE] at (.125,3) (g2a) {\textcolor{black!80!white!80}{$G^{(2)}$}};


\node at (2.75,1.5)[rectangle,rounded corners,draw=black!80!white!80,fill=black!90!white!60,minimum size=8mm] (c6){\textcolor{white}{}};
\node at (4.25,1.5)[rectangle,rounded corners,draw=black!80!white!80,fill=black!90!white!60,minimum size=8mm] (c7){\textcolor{white}{}};
\node at (5.75,1.5)[rectangle,rounded corners,draw=black!80!white!80,fill=black!90!white!60,minimum size=8mm] (c8){\textcolor{white}{}};

\node at (0,0)[yshift=-1.5cm,xshift=2.5cm,circle,draw=black!40!red!90,fill=black!10!red!90,minimum size=11mm,font=\footnotesize] (p6) {};
\node at (0,0)[yshift=-1.5cm,xshift=4.2cm,circle,draw=black!40!orange!90,fill=black!10!orange!90,minimum size=11mm] (p7) {};
\node at (0,0)[yshift=-1.5cm,xshift=5.9cm,circle,draw=black!40!orange!90,fill=black!10!orange!90,minimum size=11mm] (p8) {};

\node[font=\LARGE] at (4.25,3) (g3a) {\textcolor{black!80!white!80}{$G^{(3)}$}};
\draw[brace_nonmirror,color=black!80!white!80] (2.25,1.5)--(6.25,1.5);


\draw[line,black!80!white!80] (p1.north)--(c1.south);
\draw[line,black!80!white!80] (p1.north)--(c2.south);
\draw[line,black!80!white!80] (p2.north)--(c1.south);
\draw[line,black!80!white!80] (p2.north)--(c3.south);
\draw[line,black!80!white!80] (p3.north)--(c1.south);
\draw[line,black!80!white!80] (p3.north)--(c3.south);
\draw[line,black!80!white!80] (p3.north)--(c4.south);
\draw[line,black!80!white!80] (p4.north)--(c4.south);
\draw[line,black!80!white!80] (p4.north)--(c5.south);
\draw[line,black!80!white!80] (p5.north)--(c4.south);
\draw[line,black!80!white!80] (p5.north)--(c5.south);
\draw[line,black!80!white!80] (p6.north)--(c5.south);
\draw[line,black!80!white!80] (p6.north)--(c6.south);
\draw[line,black!80!white!80] (p7.north)--(c6.south);
\draw[line,black!80!white!80] (p7.north)--(c7.south);
\draw[line,black!80!white!80] (p7.north)--(c8.south);
\draw[line,black!80!white!80] (p8.north)--(c7.south);
\draw[line,black!80!white!80] (p8.north)--(c8.south);

\end{tikzpicture}
\caption{Step 1: cluster $1$ fails.}
\end{subfigure}\\
\begin{subfigure}[b]{.5\textwidth}
\begin{tikzpicture}[scale=.5, transform shape]

\node at (-5.75,1.5)[rectangle,rounded corners,draw=black!80!white!80,fill=black!90!white!60,minimum size=8mm] (c1){\textcolor{white}{}};
\node at (-4.25,1.5)[rectangle,rounded corners,draw=black!80!white!80,fill=black!90!white!60,minimum size=8mm] (c2){\textcolor{white}{}};
\node at (-2.75,1.5)[rectangle,rounded corners,draw=black!80!white!80,fill=black!90!white!60,minimum size=8mm] (c3){\textcolor{white}{}};

\node at (-6,-1.5)[circle,draw=black!40!orange!90,fill=black!10!orange!90,minimum size=11mm] (p1) {};
\node at (-4.3,-1.5)[circle,draw=black!40!red!90,fill=black!10!red!90,minimum size=11mm] (p2) {};
\node at (-2.6,-1.5)[circle,draw=black!40!red!90,fill=black!10!red!90,minimum size=11mm,font=\footnotesize] (p3) {$$};

\node[font=\LARGE] at (-4.25,3) (g1a) {\textcolor{black!80!white!80}{$G^{(1)}$}};
\draw[brace_nonmirror,color=black!80!white!80] (-6.25,1.5)--(-2.25,1.5);


\node at (-.65,1.5)[rectangle,rounded corners,draw=black!80!white!80,fill=black!90!white!60,minimum size=8mm] (c4) {\textcolor{white}{}};
\node at (0.9,1.5)[rectangle,rounded corners,draw=black!80!white!80,fill=black!90!white!60,minimum size=8mm] (c5) {\textcolor{white}{}};

\node at (-.9,-1.5)[circle,draw=black!40!orange!90,fill=black!10!orange!90,minimum size=11mm,font=\footnotesize] (p4) {};
\node at (.8,-1.5)[circle,draw=black!40!orange!90,fill=black!10!orange!90,minimum size=11mm,font=\footnotesize] (p5) {};

\draw[brace_nonmirror,color=black!80!white!80] (-1.15,1.5)--(1.4,1.5);
\node [font=\LARGE] at (.125,3) (g2a) {\textcolor{black!80!white!80}{$G^{(2)}$}};

\node [trapezium, trapezium angle=68, draw,black!80!white!80,inner xsep=0pt,outer sep=0pt,rounded corners, minimum height=5.75 cm, dashed, text width=1.25mm] at (0,0.6) {};

\node at (2.75,1.5)[rectangle,rounded corners,draw=black!80!white!80,fill=black!90!white!60,minimum size=8mm] (c6){\textcolor{white}{}};
\node at (4.25,1.5)[rectangle,rounded corners,draw=black!80!white!80,fill=black!90!white!60,minimum size=8mm] (c7){\textcolor{white}{}};
\node at (5.75,1.5)[rectangle,rounded corners,draw=black!80!white!80,fill=black!90!white!60,minimum size=8mm] (c8){\textcolor{white}{}};

\node at (0,0)[yshift=-1.5cm,xshift=2.5cm,circle,draw=black!40!red!90,fill=black!10!red!90,minimum size=11mm,font=\footnotesize] (p6) {};
\node at (0,0)[yshift=-1.5cm,xshift=4.2cm,circle,draw=black!40!orange!90,fill=black!10!orange!90,minimum size=11mm] (p7) {};
\node at (0,0)[yshift=-1.5cm,xshift=5.9cm,circle,draw=black!40!orange!90,fill=black!10!orange!90,minimum size=11mm] (p8) {};

\node[font=\LARGE] at (4.25,3) (g3a) {\textcolor{black!80!white!80}{$G^{(3)}$}};
\draw[brace_nonmirror,color=black!80!white!80] (2.25,1.5)--(6.25,1.5);


\draw[line,black!80!white!80] (p1.north)--(c1.south);
\draw[line,black!80!white!80] (p1.north)--(c2.south);
\draw[line,black!80!white!80] (p2.north)--(c1.south);
\draw[line,black!80!white!80] (p2.north)--(c3.south);
\draw[line,black!80!white!80] (p3.north)--(c1.south);
\draw[line,black!80!white!80] (p3.north)--(c3.south);
\draw[line,black!80!white!80] (p3.north)--(c4.south);
\draw[line,black!80!white!80] (p4.north)--(c4.south);
\draw[line,black!80!white!80] (p4.north)--(c5.south);
\draw[line,black!80!white!80] (p5.north)--(c4.south);
\draw[line,black!80!white!80] (p5.north)--(c5.south);
\draw[line,black!80!white!80] (p6.north)--(c5.south);
\draw[line,black!80!white!80] (p6.north)--(c6.south);
\draw[line,black!80!white!80] (p7.north)--(c6.south);
\draw[line,black!80!white!80] (p7.north)--(c7.south);
\draw[line,black!80!white!80] (p7.north)--(c8.south);
\draw[line,black!80!white!80] (p8.north)--(c7.south);
\draw[line,black!80!white!80] (p8.north)--(c8.south);

\end{tikzpicture}
\subcaption{Step 2: cluster $2$ fails.}
\end{subfigure}\quad
\begin{subfigure}[b]{.5\textwidth}
\begin{tikzpicture}[scale=.5, transform shape]

\node at (-5.75,1.5)[rectangle,rounded corners,draw=black!80!white!80,fill=black!90!white!60,minimum size=8mm] (c1){\textcolor{white}{}};
\node at (-4.25,1.5)[rectangle,rounded corners,draw=black!80!white!80,fill=black!90!white!60,minimum size=8mm] (c2){\textcolor{white}{}};
\node at (-2.75,1.5)[rectangle,rounded corners,draw=black!80!white!80,fill=black!90!white!60,minimum size=8mm] (c3){\textcolor{white}{}};

\node at (-6,-1.5)[circle,draw=black!40!orange!90,fill=black!10!orange!90,minimum size=11mm] (p1) {};
\node at (-4.3,-1.5)[circle,draw=black!40!red!90,fill=black!10!red!90,minimum size=11mm] (p2) {};
\node at (-2.6,-1.5)[circle,draw=black!40!red!90,fill=black!10!red!90,minimum size=11mm,font=\footnotesize] (p3) {$$};

\node[font=\LARGE] at (-4.25,3) (g1a) {\textcolor{black!80!white!80}{$G^{(1)}$}};
\draw[brace_nonmirror,color=black!80!white!80] (-6.25,1.5)--(-2.25,1.5);


\node at (-.65,1.5)[rectangle,rounded corners,draw=black!80!white!80,fill=black!90!white!60,minimum size=8mm] (c4) {\textcolor{white}{}};
\node at (0.9,1.5)[rectangle,rounded corners,draw=black!80!white!80,fill=black!90!white!60,minimum size=8mm] (c5) {\textcolor{white}{}};

\node at (-.9,-1.5)[circle,draw=black!40!orange!90,fill=black!10!orange!90,minimum size=11mm,font=\footnotesize] (p4) {};
\node at (.8,-1.5)[circle,draw=black!40!orange!90,fill=black!10!orange!90,minimum size=11mm,font=\footnotesize] (p5) {};

\draw[brace_nonmirror,color=black!80!white!80] (-1.15,1.5)--(1.4,1.5);
\node [font=\LARGE] at (.125,3) (g2a) {\textcolor{black!80!white!80}{$G^{(2)}$}};


\node at (2.75,1.5)[rectangle,rounded corners,draw=black!80!white!80,fill=black!90!white!60,minimum size=8mm] (c6){\textcolor{white}{}};
\node at (4.25,1.5)[rectangle,rounded corners,draw=black!80!white!80,fill=black!90!white!60,minimum size=8mm] (c7){\textcolor{white}{}};
\node at (5.75,1.5)[rectangle,rounded corners,draw=black!80!white!80,fill=black!90!white!60,minimum size=8mm] (c8){\textcolor{white}{}};

\node at (0,0)[yshift=-1.5cm,xshift=2.5cm,circle,draw=black!40!green!80,fill=black!30!green!80,minimum size=11mm,font=\footnotesize] (p6) {};
\node at (0,0)[yshift=-1.5cm,xshift=4.2cm,circle,draw=black!40!orange!90,fill=black!10!orange!90,minimum size=11mm] (p7) {};
\node at (0,0)[yshift=-1.5cm,xshift=5.9cm,circle,draw=black!40!orange!90,fill=black!10!orange!90,minimum size=11mm] (p8) {};

\node[font=\LARGE] at (4.25,3) (g3a) {\textcolor{black!80!white!80}{$G^{(3)}$}};
\draw[brace_nonmirror,color=black!80!white!80] (2.25,1.5)--(6.25,1.5);

\draw[dashed,black!80!white!80,rounded corners] (1.6,3.5) rectangle (6.75,-2.25);

\draw[line,black!80!white!80] (p1.north)--(c1.south);
\draw[line,black!80!white!80] (p1.north)--(c2.south);
\draw[line,black!80!white!80] (p2.north)--(c1.south);
\draw[line,black!80!white!80] (p2.north)--(c3.south);
\draw[line,black!80!white!80] (p3.north)--(c1.south);
\draw[line,black!80!white!80] (p3.north)--(c3.south);
\draw[line,black!80!white!80] (p3.north)--(c4.south);
\draw[line,black!80!white!80] (p4.north)--(c4.south);
\draw[line,black!80!white!80] (p4.north)--(c5.south);
\draw[line,black!80!white!80] (p5.north)--(c4.south);
\draw[line,black!80!white!80] (p5.north)--(c5.south);
\draw[line,black!80!white!80] (p6.north)--(c5.south);
\draw[line,black!80!white!80] (p6.north)--(c6.south);
\draw[line,black!80!white!80] (p7.north)--(c6.south);
\draw[line,black!80!white!80] (p7.north)--(c7.south);
\draw[line,black!80!white!80] (p7.north)--(c8.south);
\draw[line,black!80!white!80] (p8.north)--(c7.south);
\draw[line,black!80!white!80] (p8.north)--(c8.south);

\end{tikzpicture}
\caption{Step 3: cluster $3$ succeeds.}
\end{subfigure}\\
\begin{subfigure}[b]{.5\textwidth}
\begin{tikzpicture}[scale=.5, transform shape]

\node at (-5.75,1.5)[rectangle,rounded corners,draw=black!80!white!80,fill=black!90!white!60,minimum size=8mm] (c1){\textcolor{white}{}};
\node at (-4.25,1.5)[rectangle,rounded corners,draw=black!80!white!80,fill=black!90!white!60,minimum size=8mm] (c2){\textcolor{white}{}};
\node at (-2.75,1.5)[rectangle,rounded corners,draw=black!80!white!80,fill=black!90!white!60,minimum size=8mm] (c3){\textcolor{white}{}};

\node at (-6,-1.5)[circle,draw=black!40!orange!90,fill=black!10!orange!90,minimum size=11mm] (p1) {};
\node at (-4.3,-1.5)[circle,draw=black!40!red!90,fill=black!10!red!90,minimum size=11mm] (p2) {};
\node at (-2.6,-1.5)[circle,draw=black!40!red!90,fill=black!10!red!90,minimum size=11mm,font=\footnotesize] (p3) {$$};

\node[font=\LARGE] at (-4.25,3) (g1a) {\textcolor{black!80!white!80}{$G^{(1)}$}};
\draw[brace_nonmirror,color=black!80!white!80] (-6.25,1.5)--(-2.25,1.5);

\draw[dashed,black!80!white!80,rounded corners] (-6.75,3.5) rectangle (-1.75,-2.25);

\node at (-.65,1.5)[rectangle,rounded corners,draw=black!80!white!80,fill=black!90!white!60,minimum size=8mm] (c4) {\textcolor{white}{}};
\node at (0.9,1.5)[rectangle,rounded corners,draw=black!80!white!80,fill=black!90!white!60,minimum size=8mm] (c5) {\textcolor{white}{}};

\node at (-.9,-1.5)[circle,draw=black!40!orange!90,fill=black!10!orange!90,minimum size=11mm,font=\footnotesize] (p4) {};
\node at (.8,-1.5)[circle,draw=black!40!orange!90,fill=black!10!orange!90,minimum size=11mm,font=\footnotesize] (p5) {};

\draw[brace_nonmirror,color=black!80!white!80] (-1.15,1.5)--(1.4,1.5);
\node [font=\LARGE] at (.125,3) (g2a) {\textcolor{black!80!white!80}{$G^{(2)}$}};


\node at (2.75,1.5)[rectangle,rounded corners,draw=black!80!white!80,fill=black!90!white!60,minimum size=8mm] (c6){\textcolor{white}{}};
\node at (4.25,1.5)[rectangle,rounded corners,draw=black!80!white!80,fill=black!90!white!60,minimum size=8mm] (c7){\textcolor{white}{}};
\node at (5.75,1.5)[rectangle,rounded corners,draw=black!80!white!80,fill=black!90!white!60,minimum size=8mm] (c8){\textcolor{white}{}};

\node at (0,0)[yshift=-1.5cm,xshift=2.5cm,circle,draw=black!40!orange!90,fill=black!10!orange!90,minimum size=11mm,font=\footnotesize] (p6) {};
\node at (0,0)[yshift=-1.5cm,xshift=4.2cm,circle,draw=black!40!orange!90,fill=black!10!orange!90,minimum size=11mm] (p7) {};
\node at (0,0)[yshift=-1.5cm,xshift=5.9cm,circle,draw=black!40!orange!90,fill=black!10!orange!90,minimum size=11mm] (p8) {};

\node[font=\LARGE] at (4.25,3) (g3a) {\textcolor{black!80!white!80}{$G^{(3)}$}};
\draw[brace_nonmirror,color=black!80!white!80] (2.25,1.5)--(6.25,1.5);


\draw[line,black!80!white!80] (p1.north)--(c1.south);
\draw[line,black!80!white!80] (p1.north)--(c2.south);
\draw[line,black!80!white!80] (p2.north)--(c1.south);
\draw[line,black!80!white!80] (p2.north)--(c3.south);
\draw[line,black!80!white!80] (p3.north)--(c1.south);
\draw[line,black!80!white!80] (p3.north)--(c3.south);
\draw[line,black!80!white!80] (p3.north)--(c4.south);
\draw[line,black!80!white!80] (p4.north)--(c4.south);
\draw[line,black!80!white!80] (p4.north)--(c5.south);
\draw[line,black!80!white!80] (p5.north)--(c4.south);
\draw[line,black!80!white!80] (p5.north)--(c5.south);
\draw[line,black!80!white!80] (p6.north)--(c5.south);
\draw[line,black!80!white!80] (p6.north)--(c6.south);
\draw[line,black!80!white!80] (p7.north)--(c6.south);
\draw[line,black!80!white!80] (p7.north)--(c7.south);
\draw[line,black!80!white!80] (p7.north)--(c8.south);
\draw[line,black!80!white!80] (p8.north)--(c7.south);
\draw[line,black!80!white!80] (p8.north)--(c8.south);

\end{tikzpicture}
\subcaption{Step 4: cluster $1$ fails again.}
\end{subfigure}\quad
\begin{subfigure}[b]{.5\textwidth}
\begin{tikzpicture}[scale=.5, transform shape]

\node at (-5.75,1.5)[rectangle,rounded corners,draw=black!80!white!80,fill=black!90!white!60,minimum size=8mm] (c1){\textcolor{white}{}};
\node at (-4.25,1.5)[rectangle,rounded corners,draw=black!80!white!80,fill=black!90!white!60,minimum size=8mm] (c2){\textcolor{white}{}};
\node at (-2.75,1.5)[rectangle,rounded corners,draw=black!80!white!80,fill=black!90!white!60,minimum size=8mm] (c3){\textcolor{white}{}};

\node at (-6,-1.5)[circle,draw=black!40!orange!90,fill=black!10!orange!90,minimum size=11mm] (p1) {};
\node at (-4.3,-1.5)[circle,draw=black!40!red!90,fill=black!10!red!90,minimum size=11mm] (p2) {};
\node at (-2.6,-1.5)[circle,draw=black!40!green!80,fill=black!30!green!80,minimum size=11mm,font=\footnotesize] (p3) {$$};

\node[font=\LARGE] at (-4.25,3) (g1a) {\textcolor{black!80!white!80}{$G^{(1)}$}};
\draw[brace_nonmirror,color=black!80!white!80] (-6.25,1.5)--(-2.25,1.5);


\node at (-.65,1.5)[rectangle,rounded corners,draw=black!80!white!80,fill=black!90!white!60,minimum size=8mm] (c4) {\textcolor{white}{}};
\node at (0.9,1.5)[rectangle,rounded corners,draw=black!80!white!80,fill=black!90!white!60,minimum size=8mm] (c5) {\textcolor{white}{}};

\node at (-.9,-1.5)[circle,draw=black!40!orange!90,fill=black!10!orange!90,minimum size=11mm,font=\footnotesize] (p4) {};
\node at (.8,-1.5)[circle,draw=black!40!orange!90,fill=black!10!orange!90,minimum size=11mm,font=\footnotesize] (p5) {};

\draw[brace_nonmirror,color=black!80!white!80] (-1.15,1.5)--(1.4,1.5);
\node [font=\LARGE] at (.125,3) (g2a) {\textcolor{black!80!white!80}{$G^{(2)}$}};

\node [trapezium, trapezium angle=68, draw,black!80!white!80,inner xsep=0pt,outer sep=0pt,rounded corners,
  minimum height=5.75 cm, dashed, text width=1.25mm] at (0,0.6) {};

\node at (2.75,1.5)[rectangle,rounded corners,draw=black!80!white!80,fill=black!90!white!60,minimum size=8mm] (c6){\textcolor{white}{}};
\node at (4.25,1.5)[rectangle,rounded corners,draw=black!80!white!80,fill=black!90!white!60,minimum size=8mm] (c7){\textcolor{white}{}};
\node at (5.75,1.5)[rectangle,rounded corners,draw=black!80!white!80,fill=black!90!white!60,minimum size=8mm] (c8){\textcolor{white}{}};

\node at (0,0)[yshift=-1.5cm,xshift=2.5cm,circle,draw=black!40!orange!90,fill=black!10!orange!90,minimum size=11mm,font=\footnotesize] (p6) {};
\node at (0,0)[yshift=-1.5cm,xshift=4.2cm,circle,draw=black!40!orange!90,fill=black!10!orange!90,minimum size=11mm] (p7) {};
\node at (0,0)[yshift=-1.5cm,xshift=5.9cm,circle,draw=black!40!orange!90,fill=black!10!orange!90,minimum size=11mm] (p8) {};

\node[font=\LARGE] at (4.25,3) (g3a) {\textcolor{black!80!white!80}{$G^{(3)}$}};
\draw[brace_nonmirror,color=black!80!white!80] (2.25,1.5)--(6.25,1.5);


\draw[line,black!80!white!80] (p1.north)--(c1.south);
\draw[line,black!80!white!80] (p1.north)--(c2.south);
\draw[line,black!80!white!80] (p2.north)--(c1.south);
\draw[line,black!80!white!80] (p2.north)--(c3.south);
\draw[line,black!80!white!80] (p3.north)--(c1.south);
\draw[line,black!80!white!80] (p3.north)--(c3.south);
\draw[line,black!80!white!80] (p3.north)--(c4.south);
\draw[line,black!80!white!80] (p4.north)--(c4.south);
\draw[line,black!80!white!80] (p4.north)--(c5.south);
\draw[line,black!80!white!80] (p5.north)--(c4.south);
\draw[line,black!80!white!80] (p5.north)--(c5.south);
\draw[line,black!80!white!80] (p6.north)--(c5.south);
\draw[line,black!80!white!80] (p6.north)--(c6.south);
\draw[line,black!80!white!80] (p7.north)--(c6.south);
\draw[line,black!80!white!80] (p7.north)--(c7.south);
\draw[line,black!80!white!80] (p7.north)--(c8.south);
\draw[line,black!80!white!80] (p8.north)--(c7.south);
\draw[line,black!80!white!80] (p8.north)--(c8.south);

\end{tikzpicture}
\subcaption{Step 5: cluster $2$ succeeds.}
\end{subfigure}\\
\begin{subfigure}[b]{.5\textwidth}
\begin{tikzpicture}[scale=.5, transform shape]

\node at (-5.75,1.5)[rectangle,rounded corners,draw=black!80!white!80,fill=black!90!white!60,minimum size=8mm] (c1){\textcolor{white}{}};
\node at (-4.25,1.5)[rectangle,rounded corners,draw=black!80!white!80,fill=black!90!white!60,minimum size=8mm] (c2){\textcolor{white}{}};
\node at (-2.75,1.5)[rectangle,rounded corners,draw=black!80!white!80,fill=black!90!white!60,minimum size=8mm] (c3){\textcolor{white}{}};

\node at (-6,-1.5)[circle,draw=black!40!orange!90,fill=black!10!orange!90,minimum size=11mm] (p1) {};
\node at (-4.3,-1.5)[circle,draw=black!40!green!80,fill=black!30!green!80,minimum size=11mm] (p2) {};
\node at (-2.6,-1.5)[circle,draw=black!40!orange!90,fill=black!10!orange!90,minimum size=11mm,font=\footnotesize] (p3) {$$};

\node[font=\LARGE] at (-4.25,3) (g1a) {\textcolor{black!80!white!80}{$G^{(1)}$}};
\draw[brace_nonmirror,color=black!80!white!80] (-6.25,1.5)--(-2.25,1.5);

\draw[dashed,black!80!white!80,rounded corners] (-6.75,3.5) rectangle (-1.75,-2.25);

\node at (-.65,1.5)[rectangle,rounded corners,draw=black!80!white!80,fill=black!90!white!60,minimum size=8mm] (c4) {\textcolor{white}{}};
\node at (0.9,1.5)[rectangle,rounded corners,draw=black!80!white!80,fill=black!90!white!60,minimum size=8mm] (c5) {\textcolor{white}{}};

\node at (-.9,-1.5)[circle,draw=black!40!orange!90,fill=black!10!orange!90,minimum size=11mm,font=\footnotesize] (p4) {};
\node at (.8,-1.5)[circle,draw=black!40!orange!90,fill=black!10!orange!90,minimum size=11mm,font=\footnotesize] (p5) {};

\draw[brace_nonmirror,color=black!80!white!80] (-1.15,1.5)--(1.4,1.5);
\node [font=\LARGE] at (.125,3) (g2a) {\textcolor{black!80!white!80}{$G^{(2)}$}};


\node at (2.75,1.5)[rectangle,rounded corners,draw=black!80!white!80,fill=black!90!white!60,minimum size=8mm] (c6){\textcolor{white}{}};
\node at (4.25,1.5)[rectangle,rounded corners,draw=black!80!white!80,fill=black!90!white!60,minimum size=8mm] (c7){\textcolor{white}{}};
\node at (5.75,1.5)[rectangle,rounded corners,draw=black!80!white!80,fill=black!90!white!60,minimum size=8mm] (c8){\textcolor{white}{}};

\node at (0,0)[yshift=-1.5cm,xshift=2.5cm,circle,draw=black!40!orange!90,fill=black!10!orange!90,minimum size=11mm,font=\footnotesize] (p6) {};
\node at (0,0)[yshift=-1.5cm,xshift=4.2cm,circle,draw=black!40!orange!90,fill=black!10!orange!90,minimum size=11mm] (p7) {};
\node at (0,0)[yshift=-1.5cm,xshift=5.9cm,circle,draw=black!40!orange!90,fill=black!10!orange!90,minimum size=11mm] (p8) {};

\node[font=\LARGE] at (4.25,3) (g3a) {\textcolor{black!80!white!80}{$G^{(3)}$}};
\draw[brace_nonmirror,color=black!80!white!80] (2.25,1.5)--(6.25,1.5);


\draw[line,black!80!white!80] (p1.north)--(c1.south);
\draw[line,black!80!white!80] (p1.north)--(c2.south);
\draw[line,black!80!white!80] (p2.north)--(c1.south);
\draw[line,black!80!white!80] (p2.north)--(c3.south);
\draw[line,black!80!white!80] (p3.north)--(c1.south);
\draw[line,black!80!white!80] (p3.north)--(c3.south);
\draw[line,black!80!white!80] (p3.north)--(c4.south);
\draw[line,black!80!white!80] (p4.north)--(c4.south);
\draw[line,black!80!white!80] (p4.north)--(c5.south);
\draw[line,black!80!white!80] (p5.north)--(c4.south);
\draw[line,black!80!white!80] (p5.north)--(c5.south);
\draw[line,black!80!white!80] (p6.north)--(c5.south);
\draw[line,black!80!white!80] (p6.north)--(c6.south);
\draw[line,black!80!white!80] (p7.north)--(c6.south);
\draw[line,black!80!white!80] (p7.north)--(c7.south);
\draw[line,black!80!white!80] (p7.north)--(c8.south);
\draw[line,black!80!white!80] (p8.north)--(c7.south);
\draw[line,black!80!white!80] (p8.north)--(c8.south);

\end{tikzpicture}
\subcaption{Step 7: cluster $1$ succeeds.}
\end{subfigure}\quad
\begin{subfigure}[b]{.5\textwidth}
\begin{tikzpicture}[scale=.5, transform shape]

\node at (-5.75,1.5)[rectangle,rounded corners,draw=black!80!white!80,fill=black!90!white!60,minimum size=8mm] (c1){\textcolor{white}{}};
\node at (-4.25,1.5)[rectangle,rounded corners,draw=black!80!white!80,fill=black!90!white!60,minimum size=8mm] (c2){\textcolor{white}{}};
\node at (-2.75,1.5)[rectangle,rounded corners,draw=black!80!white!80,fill=black!90!white!60,minimum size=8mm] (c3){\textcolor{white}{}};

\node at (-6,-1.5)[circle,draw=black!40!orange!90,fill=black!10!orange!90,minimum size=11mm] (p1) {};
\node at (-4.3,-1.5)[circle,draw=black!40!orange!90,fill=black!10!orange!90,minimum size=11mm] (p2) {};
\node at (-2.6,-1.5)[circle,draw=black!40!orange!90,fill=black!10!orange!90,minimum size=11mm,font=\footnotesize] (p3) {$$};

\node[font=\LARGE] at (-4.25,3) (g1a) {\textcolor{black!80!white!80}{$G^{(1)}$}};
\draw[brace_nonmirror,color=black!80!white!80] (-6.25,1.5)--(-2.25,1.5);


\node at (-.65,1.5)[rectangle,rounded corners,draw=black!80!white!80,fill=black!90!white!60,minimum size=8mm] (c4) {\textcolor{white}{}};
\node at (0.9,1.5)[rectangle,rounded corners,draw=black!80!white!80,fill=black!90!white!60,minimum size=8mm] (c5) {\textcolor{white}{}};

\node at (-.9,-1.5)[circle,draw=black!40!orange!90,fill=black!10!orange!90,minimum size=11mm,font=\footnotesize] (p4) {};
\node at (.8,-1.5)[circle,draw=black!40!orange!90,fill=black!10!orange!90,minimum size=11mm,font=\footnotesize] (p5) {};

\draw[brace_nonmirror,color=black!80!white!80] (-1.15,1.5)--(1.4,1.5);
\node [font=\LARGE] at (.125,3) (g2a) {\textcolor{black!80!white!80}{$G^{(2)}$}};

\node [trapezium, trapezium angle=68, draw,black!80!white!80,inner xsep=0pt,outer sep=0pt,rounded corners,
  minimum height=5.75 cm, dashed, text width=1.25mm] at (0,0.6) {};

\node at (2.75,1.5)[rectangle,rounded corners,draw=black!80!white!80,fill=black!90!white!60,minimum size=8mm] (c6){\textcolor{white}{}};
\node at (4.25,1.5)[rectangle,rounded corners,draw=black!80!white!80,fill=black!90!white!60,minimum size=8mm] (c7){\textcolor{white}{}};
\node at (5.75,1.5)[rectangle,rounded corners,draw=black!80!white!80,fill=black!90!white!60,minimum size=8mm] (c8){\textcolor{white}{}};

\node at (0,0)[yshift=-1.5cm,xshift=2.5cm,circle,draw=black!40!orange!90,fill=black!10!orange!90,minimum size=11mm,font=\footnotesize] (p6) {};
\node at (0,0)[yshift=-1.5cm,xshift=4.2cm,circle,draw=black!40!orange!90,fill=black!10!orange!90,minimum size=11mm] (p7) {};
\node at (0,0)[yshift=-1.5cm,xshift=5.9cm,circle,draw=black!40!orange!90,fill=black!10!orange!90,minimum size=11mm] (p8) {};

\node[font=\LARGE] at (4.25,3) (g3a) {\textcolor{black!80!white!80}{$G^{(3)}$}};
\draw[brace_nonmirror,color=black!80!white!80] (2.25,1.5)--(6.25,1.5);


\draw[line,black!80!white!80] (p1.north)--(c1.south);
\draw[line,black!80!white!80] (p1.north)--(c2.south);
\draw[line,black!80!white!80] (p2.north)--(c1.south);
\draw[line,black!80!white!80] (p2.north)--(c3.south);
\draw[line,black!80!white!80] (p3.north)--(c1.south);
\draw[line,black!80!white!80] (p3.north)--(c3.south);
\draw[line,black!80!white!80] (p3.north)--(c4.south);
\draw[line,black!80!white!80] (p4.north)--(c4.south);
\draw[line,black!80!white!80] (p4.north)--(c5.south);
\draw[line,black!80!white!80] (p5.north)--(c4.south);
\draw[line,black!80!white!80] (p5.north)--(c5.south);
\draw[line,black!80!white!80] (p6.north)--(c5.south);
\draw[line,black!80!white!80] (p6.north)--(c6.south);
\draw[line,black!80!white!80] (p7.north)--(c6.south);
\draw[line,black!80!white!80] (p7.north)--(c7.south);
\draw[line,black!80!white!80] (p7.north)--(c8.south);
\draw[line,black!80!white!80] (p8.north)--(c7.south);
\draw[line,black!80!white!80] (p8.north)--(c8.south);

\end{tikzpicture}
\subcaption{Step 8: Algorithm finishes successfully.}
\end{subfigure}

\caption{How overlaps among clusters help the neural network to achieve better error correction performance. We assume that each cluster can correct one input error. In other words, if the number of input errors are higher than one the cluster declares a failure. \label{fig:overlapping_helps}}
\end{figure}
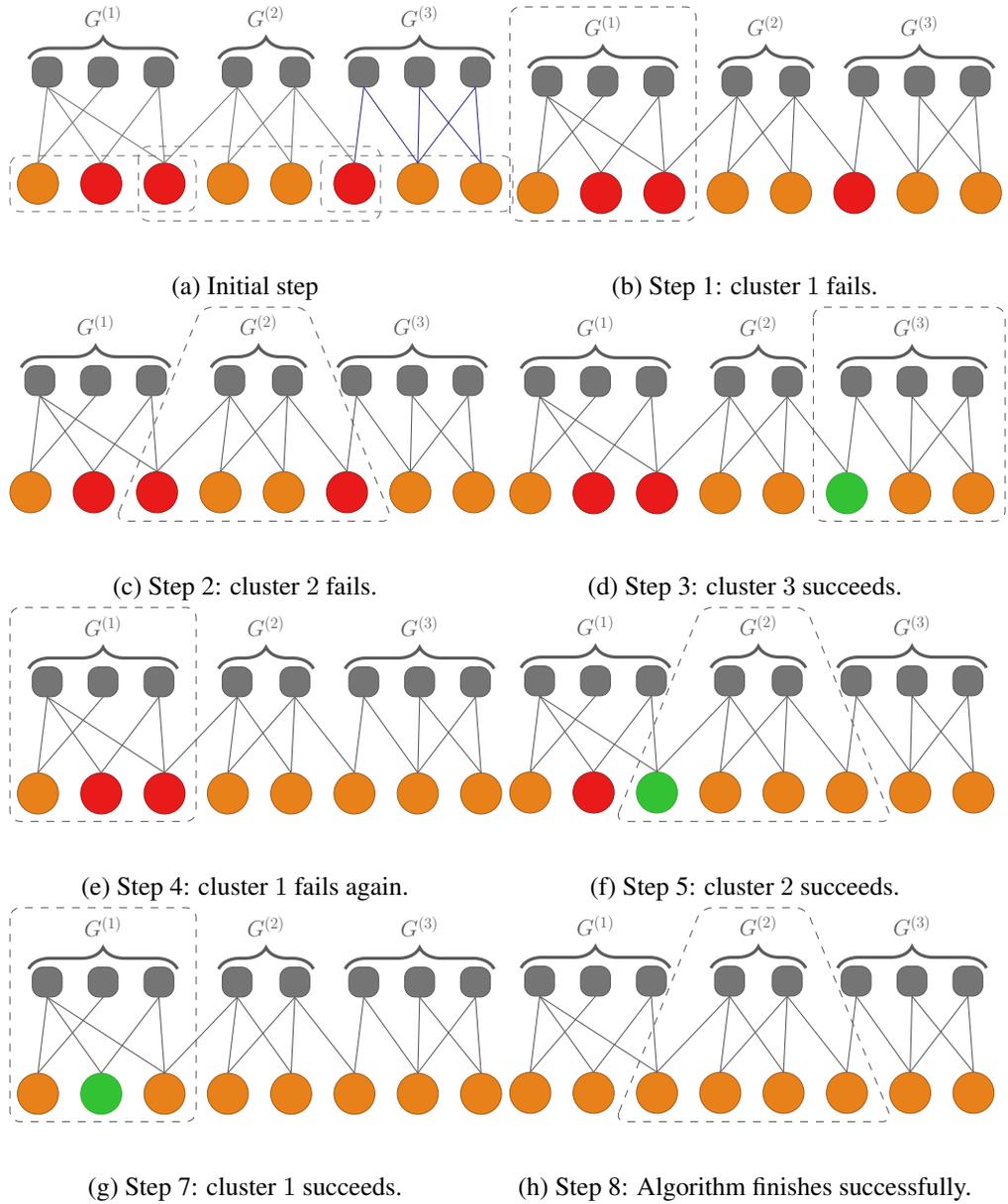

As mentioned earlier, the error correction ability of Algorithm~\ref{algo:correction} is fairly limited. As a result, if clusters work independently, they cannot correct more than a few external errors. However, as clusters overlap their combined performance can potentially be much better. Basically, they can help each other in resolving external errors: a cluster whose pattern neurons are in their correct states can provide truthful information 
to neighboring clusters. Figure~\ref{fig:overlapping_helps} illustrates this idea.
%

This property is exploited in the inter-cluster recall approach, formally given by Algorithm~\ref{algo:peeling}. In words, the inter-cluster approach proceeds by applying Algorithm~\ref{algo:correction} in a round-robin fashion to each cluster. Clusters either eliminate their internal noise in which case they keep their 
new states and can now help other clusters, or revert back to their original states. Note that by such a scheduling scheme, neurons can only change their states towards correct values. 

\begin{algorithm}[t]
\caption{Sequential Peeling Algorithm}
\label{algo:peeling}
\begin{algorithmic}[1]
\REQUIRE{$\widetilde{G}, G^{(1)}, G^{(2)}, \dots, G^{(L)}$.}
\ENSURE{$x_1,x_2,\dots,x_n$}
\WHILE{There is an unsatisfied $v^{(\ell)}$, for $\ell=1,\dots,L$}
\FOR{$\ell = 1 \to L$} 
\STATE If $v^{(\ell)}$ is unsatisfied, apply Algorithm~\ref{algo:correction} to cluster $G^{(l)}$.
\STATE If $v^{(\ell)}$ remained unsatisfied, revert the state of pattern neurons connected to $v^{(\ell)}$ to their initial state. Otherwise, keep their current states.
\ENDFOR
\ENDWHILE
\STATE Declare {$x_1,x_2,\dots,x_n$} if all $v^{(\ell)}$'s are satisfied. Otherwise, declare failure. 
\end{algorithmic}
\end{algorithm}

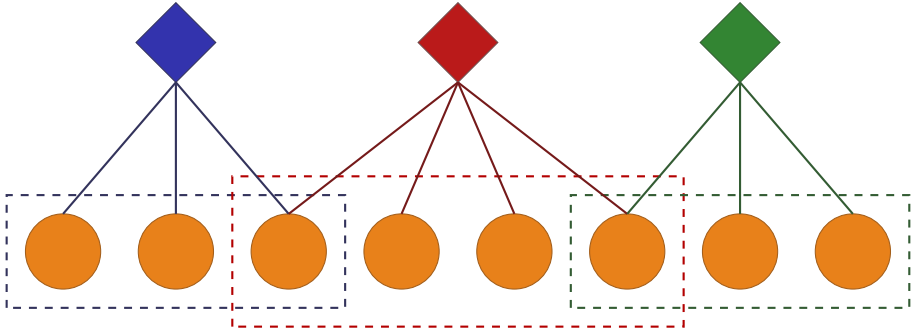
\begin{figure}[t]
\centering
\begin{tikzpicture}[scale=.5]

\draw[draw=black!80!blue!80,fill=black!40!blue!80,rotate around={45:(3,4.5)}] (3,4.5) rectangle (4.5,6);
\draw[draw=black!80!red!70,fill=black!30!red!90,rotate around={45:(10.5,4.5)}] (10.5,4.5) rectangle (12,6);
\draw[draw=black!80!green!80,fill=black!60!green!80,rotate around={45:(18,4.5)}] (18,4.5) rectangle (19.5,6);

\draw[draw=black!40!orange!90,fill=black!10!orange!90] (0,0) circle (1cm);
\draw[draw=black!40!orange!90,fill=black!10!orange!90] (3,0) circle (1cm);
\draw[draw=black!40!orange!90,fill=black!10!orange!90] (6,0) circle (1cm);
\draw[draw=black!40!orange!90,fill=black!10!orange!90] (9,0) circle (1cm);
\draw[draw=black!40!orange!90,fill=black!10!orange!90] (12,0) circle (1cm);
\draw[draw=black!40!orange!90,fill=black!10!orange!90] (15,0) circle (1cm);
\draw[draw=black!40!orange!90,fill=black!10!orange!90] (18,0) circle (1cm);
\draw[draw=black!40!orange!90,fill=black!10!orange!90] (21,0) circle (1cm);

\draw[draw=black!80!blue!80,dashed,thick] (-1.5,-1.5) rectangle (7.5,1.5);

\draw[draw=black!30!red!100,dashed,thick] (4.5,-2) rectangle (16.5,2);

\draw[draw=black!80!green!80,dashed,thick] (13.5,-1.5) rectangle (22.5,1.5);

\draw[thick,black!80!blue!80] (0,1) --(3,4.5);
\draw[thick,black!80!blue!80] (3,1) --(3,4.5);
\draw[thick,black!80!blue!80] (6,1) --(3,4.5);

\draw[thick,black!60!red!90] (6,1) -- (10.5,4.5);
\draw[thick,black!60!red!90] (9,1) -- (10.5,4.5);
\draw[thick,black!60!red!90] (12,1) -- (10.5,4.5);
\draw[thick,black!60!red!90] (15,1) --(10.5,4.5);

\draw[thick,black!80!green!80] (15,1) --(18,4.5);
\draw[thick,black!80!green!80] (18,1) --(18,4.5);
\draw[thick,black!80!green!80] (21,1) --(18,4.5);
\end{tikzpicture}
\caption{Contraction graph $\widetilde{G}$ corresponding to graph $G$ in Figure~\ref{fig:overlapping_clustered_network}.\label{fig:inter}}
\end{figure}

The inter-cluster algorithm is in spirit similar to a famous decoding algorithm in communication systems for erasure channels, called the \emph{peeling algorithm} \citep{erasure}. To make the connection more concrete, we first need to define a \textit{contracted} version of the neural graph $G$ as follows. 
In the contracted graph $\widetilde{G}$, we compress all constraint nodes of a cluster $G^{(\ell)}$ into a single \emph{super constraint node} $v^{(\ell)}$ (see Figure~\ref{fig:inter}). Then, each super constraint node essentially acts as a check node capable of detecting and correcting any single error among its neighbors (i.e., pattern neurons). In contrast, it declares a failure if two or more of its neighbors are corrupted by noise. Once an error is corrected by a cluster, the number of errors in overlapping clusters may also reduce which in turn help them to eliminate their errors. 


Through introducing the contracted graph, the similarity to the Peeling Decoder is now evident: in the Peeling Decoder, each constraint (so called checksum) node is capable of correcting a single erasure among its neighbors. Similarly, if there are more than one erasure among the neighbors, the checksum node declares erasure.   However, once an erasure is eliminated by a checksum node, it helps other constraint nodes, namely those connected to the recently-eliminated erased node, as they will have one less erasure among their neighbors to deal with. 

Based on the above similarity, we borrow methods from modern coding theory to obtain theoretical guarantees on the error rate of our proposed recall algorithm. More specifically, we use Density Evolution (DE), first developed by \citet{erasure} and generalized by \citet{urbanke}, to accurately bound the error correction performance. 

%

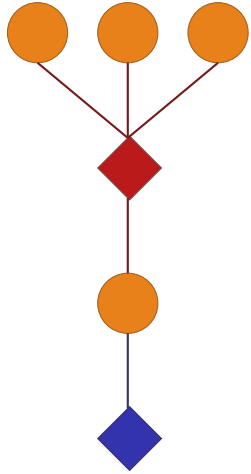
\begin{figure}[t]
\centering
\begin{tikzpicture}[scale=.4]

\draw[draw=black!80!blue!80,fill=black!40!blue!80,rotate around={45:(0,1.5)}] (0,0) rectangle (1.5,1.5);

\draw[draw=black!40!orange!90,fill=black!10!orange!90] (1,6) circle (1cm);

\draw[draw=black!80!red!70,fill=black!30!red!90,rotate around={45:(0,10.5)}] (0,9) rectangle (1.5,10.5);

\draw[draw=black!40!orange!90,fill=black!10!orange!90] (-2,15) circle (1cm);
\draw[draw=black!40!orange!90,fill=black!10!orange!90] (1,15) circle (1cm);
\draw[draw=black!40!orange!90,fill=black!10!orange!90] (4,15) circle (1cm);


\draw[thick,black!80!blue!80] (1,2.5) --(1,5);

\draw[thick,black!60!red!90] (1,7) -- (1,9.5);
\draw[thick,black!60!red!90] (1,11.5) -- (-2,14);
\draw[thick,black!60!red!90] (1,11.5) -- (1,14);
\draw[thick,black!60!red!90] (1,11.5) --(4,14);

\end{tikzpicture}
\caption{The decision subgraph of depth 2 for the third edge (from left) in Figure~\ref{fig:inter}.
\label{fig:decision_tree}}
\end{figure}

Let $\widetilde{ \lambda}_i$ (resp. $\widetilde{\rho}_j$) denote the fraction of \emph{edges} that are adjacent to pattern (resp. constraint) nodes of degree $i$ (resp. $j$). We call $\{\widetilde{\lambda}_1,\dots,\widetilde{\lambda}_L\}$ the pattern degree distribution and
$\{\widetilde{\rho}_1,\dots,\widetilde{\rho}_n\}$ the super constraint degree distribution. Similar to Section~\ref{section_recall_intra_cluster}, it is convenient to define the degree distribution polynomials
as follows:
\begin{eqnarray*}
\widetilde{\lambda}(z) &=& \sum_{i} \widetilde{\lambda}_i z^{i-1}, \\
\widetilde{\rho}(z) &=& \sum_i \rho_i z^{i-1}.
\end{eqnarray*}
Now consider a given cluster $v^{(\ell)}$ and a pattern neuron $x$ connected to $v^{(\ell)}$. The \textit{decision subgraph} of $x$ is defined as the subgraph rooted at $x$ and branched out from the super constraint nodes, excluding $v^{(\ell)}$. If the decision subgraph is a tree up to a depth of $\tau$ (meaning that no node appears more than once), we say that the \textit{tree assumption} holds for $\tau$ levels. An example of the decision subgraph is shown in Figure \ref{fig:decision_tree}. Finally, we say that the node $v^{(\ell)}$ is unsatisfied if it is connected to a noisy pattern node. Recall that $P_c$ denotes the average probability of a super constraint node correcting a single error among its neighbors.

%
%

%
%

%

\begin{theorem}\label{th:outside}
Assume that $\tilde{G}$ is chosen randomly according to the degree distribution pair $\tilde{\lambda}$ and $\tilde{\rho}$. Then, as the number of vertices of $G$ grows large, Algorithm~\ref{algo:peeling} will succeed in correcting all errors with high probability as long as $p_e \widetilde{\lambda}\left(1-P_c\widetilde{\rho}(1-z)\right)<z$ for $z\in (0,p_e)$. 
%
%
\end{theorem}
It is worth to make a few remarks about Theorem~\ref{th:outside}. First, the condition given in Theorem~\ref{th:outside} can be used to calculate the maximal fraction of errors Algorithm~\ref{algo:peeling} can correct. For instance, for the degree distribution pair $(\widetilde{\lambda}(z) = z^2, \widetilde{\rho}(z) = z^5)$, the threshold is $p^{*}_e \approx 0.429$, below which Algorithm~\ref{algo:peeling} corrects all the errors with high probability. Second, the predicted threshold by Theorem~\ref{th:outside} is based on the pessimistic assumption that a cluster can only correct a single error. Third, for a graph $\widetilde{G}$, constructed randomly according to given degree distributions $\widetilde{\lambda}$ and $\widetilde{\rho}$, as the graph size grows the decision subgraph becomes a tree with probability close to $1$. Hence, it can be shown (see, for example \citealp{urbanke}) that the recall performance for any such graphs will be \emph{concentrated} around the average case given by Theorem \ref{th:outside}. 

\section{Pattern Retrieval Capacity}\label{section_capacity}
Before discussing the the pattern retrieval capacity, we should note that the number of patterns $C$ does not have any effect on the learning or recall algorithm except for its obvious influence on the learning time. More precisely, as long as the (sub)patterns lie on a subspace, the learning Algorithm~\ref{algo_learning} yields a matrix $W$ that is orthogonal to all the patterns of the training set. Similarly, in the recall phase, algorithms~\ref{algo:correction} and \ref{algo:peeling} only need to compute $W \cdot e$ for the noise vector $e$.

Remember that the retrieval capacity is defined as the maximum number of patterns that a neural of size $n$ is able to store. Hence, in order to show that the pattern retrieval capacity of our method is exponential in $n$, we need to demonstrate that there \emph{exists} a training set $\mc{X}$ with $C$ patterns of length $n$ for which $C \propto a^{rn}$, for some $a > 1$ and $0<r<1$. 

%

%
%

\begin{theorem}\label{theorem_exponential_solution}
Let $\mc{X}$ be a $C \times n$ matrix formed by $C$ vectors of length $n$ with entries from $\mc{Q}$. Furthermore, let $k = \lfloor rn \rfloor$ for some $0<r<1$ and $k < \min_{\ell}(n_\ell)$. Then, there exists a set of vectors of size $C = \Omega(a^{ rn })$ with some $a>1$ such that $\hbox{\text{rank}}(\mc{X}) = k <n$. Moreover, Algorithm~\ref{algo_learning} can learn this set. 
\end{theorem}
The proof is by construction. This construction can be used to synthetically generate patterns that lie in a subspace. 
\section{Experimental Results}\label{sec:experiments} 
In this section we evaluate the performance of our proposed algorithms over synthetic and natural datasets \footnote{The codes used in this paper are all available online at \url{http://goo.gl/ifR14t}.}.
\subsection{Synthetic Scenario}

A systematic way to generate patterns satisfying a set of linear constraints is outlined in the proof of Theorem \ref{theorem_exponential_solution}. This proof is constructive and provides an easy way to randomly sample patterns with linear constraints.  In our simulations, we consider a neural network in which each pattern neuron is connected to approximately $5$ clusters. The number of connections should be neither too small (to ensure information propagation) nor too big (to adhere to the sparsity requirement).
%

In the learning phase, Algorithm \ref{algo_learning} is performed (in parallel) for each cluster in order to find the connectivity matrix $W$. In the recall phase (and at each round), a pattern $x$ is sampled uniformly at random from the training set. Then, each of its entries are corrupted with $\pm 1$ additive noise independently with probability $p_e$. Algorithm~\ref{algo:peeling} is subsequently used to denoise the corrupted patterns. We average out this process over many trials to calculate the error rate and compare it to the analytic bound derived in Theorem~\ref{th:outside}.
%

\subsubsection{Learning Results} 
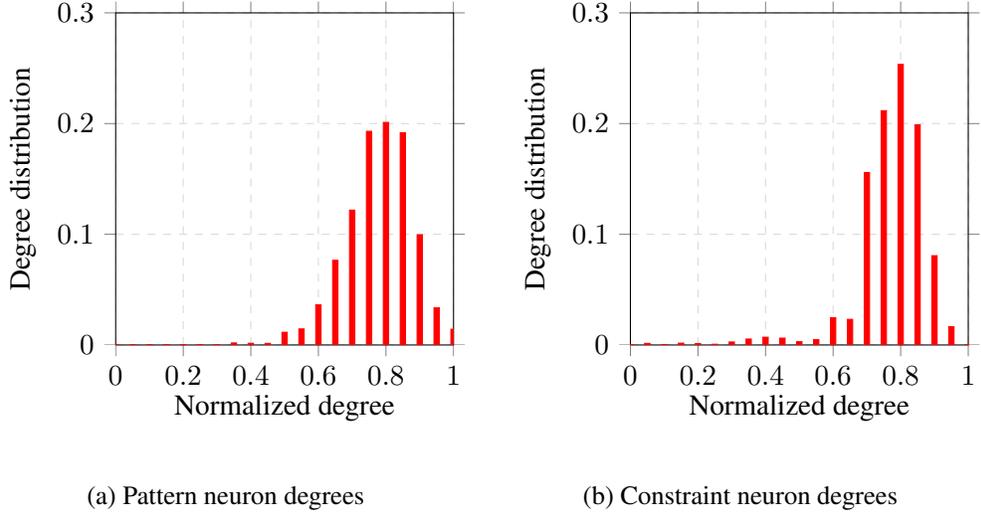
\begin{figure*}[t]%
	\begin{subfigure}[t]{0.48\textwidth}
    	\centering							  
       \centering
\begin{tikzpicture}
    \begin{axis}[
        width=\textwidth, height=6cm,     
        grid = major,
	ybar,
        grid style={dashed, gray!30},
        xmin=0,     
        xmax=1,    
        ymin=0,     
        ymax=0.3,   
        /pgfplots/xtick={0,.2,...,1}, 
	/pgfplots/ytick={0,.1,...,.4},
	bar width=2pt,
        axis background/.style={fill=white},
        ylabel=Degree distribution,
        xlabel=Normalized degree,
        tick align=outside,
	legend style={
		cells={anchor=west},
		legend pos=north east,
		}
	]
 
      \addplot [color=red,fill=red] table {Learning_Convolutional_Fig_Data_N_40_K_20_L_50_Pattern_Deg_b.dat};

   \end{axis} 
\end{tikzpicture}
		\caption{Pattern neuron degrees\label{pattern_deg_40_20_50_b}}
    \end{subfigure}\quad\quad
    \begin{subfigure}[t]{0.48\textwidth}
	    \centering						
       \centering
\begin{tikzpicture}
    \begin{axis}[
        width=\textwidth, height=6cm,     
        grid = major,
	ybar,
        grid style={dashed, gray!30},
        xmin=0,     
        xmax=1,    
        ymin=0,     
        ymax=0.3,   
        /pgfplots/xtick={0,.2,...,1}, 
	/pgfplots/ytick={0,.1,...,.4},
	bar width=2pt,
        axis background/.style={fill=white},
        ylabel=Degree distribution,
        xlabel=Normalized degree,
        tick align=outside,
	legend style={
		cells={anchor=west},
		legend pos=north east,
		}
	]
 
      \addplot [color=red,fill=red] table {Learning_Convolutional_Fig_Data_N_40_K_20_L_50_Const_Deg_b.dat};

   \end{axis} 
\end{tikzpicture}
		\caption{Constraint neuron degrees\label{const_deg_40_20_50_b}}
    \end{subfigure}
	\caption{Pattern and constraint neuron degree distributions for $n=400$, $L=50$, and an average of $20$ constraints per cluster. The learning parameters are $\alpha_t \propto 0.95/t$, $\eta = 0.75/\alpha_t$ and $\theta_t \propto 0.05/t$.\label{sparsity_inside_cluster_N_40_K_20_b}}
\end{figure*}

The left and right panels in Figure~\ref{sparsity_inside_cluster_N_40_K_20_b} illustrate the degree distributions of pattern and constraint neurons, respectively, over an ensemble of $5$ randomly generated datasets. The network size is $n=400$, which is divided into $50$ overlapping clusters, each of size around $40$, i.e., $n_\ell \simeq 40$ for $\ell=1,\dots,50$. Each pattern neuron is connected to $5$ clusters, on average. The horizontal axis shows the normalized degree of pattern (resp., constraint) neurons and the vertical axis represents the fraction of neurons with the given normalized degree. The normalization is done with respect to the number of pattern (resp., constrain) neurons in the cluster. The parameters for the learning algorithm are $\alpha_t \propto 0.95/t$, $\eta = 0.75/\alpha_t$ and $\theta_t \propto 0.05/t$.

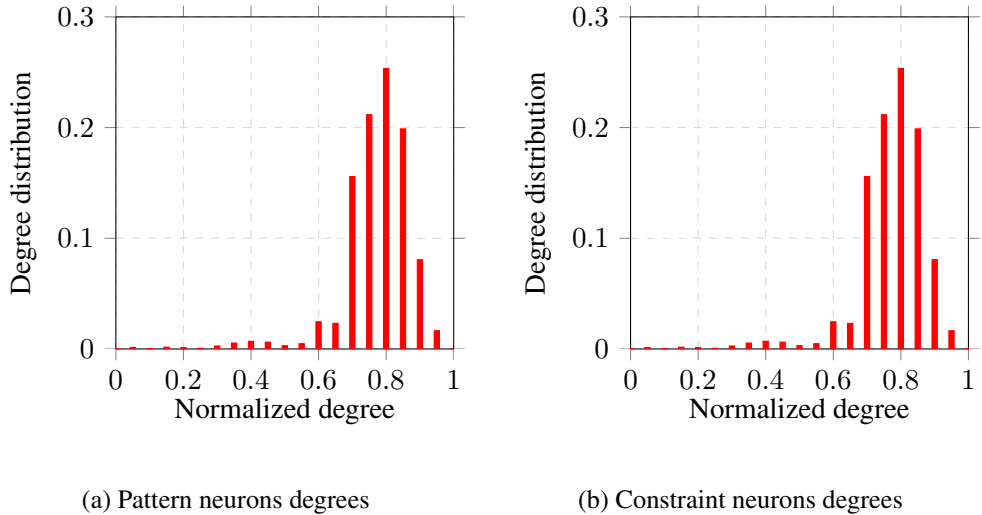
\begin{figure*}[t]%
	\begin{subfigure}[t]{0.48\textwidth}
    	\centering						
       \centering
\begin{tikzpicture}
    \begin{axis}[
        width=\textwidth, height=6cm,     
        grid = major,
	ybar,
        grid style={dashed, gray!30},
        xmin=0,     
        xmax=1,    
        ymin=0,     
        ymax=0.3,   
        /pgfplots/xtick={0,.2,...,1}, 
	/pgfplots/ytick={0,.1,...,.4},
	bar width=2pt,
        axis background/.style={fill=white},
        ylabel=Degree distribution,
        xlabel=Normalized degree,
        tick align=outside,
	legend style={
		cells={anchor=west},
		legend pos=north east,
		}
	]
 
      \addplot [color=red,fill=red] table {Learning_Convolutional_Fig_Data_N_40_K_20_L_50_Const_Deg_b.dat};

   \end{axis} 
\end{tikzpicture}
		\caption{Pattern neurons degrees\label{pattern_deg_80_40_60}}
    \end{subfigure}\quad\quad
    \begin{subfigure}[t]{0.48\textwidth}
	    \centering						
       \centering
\begin{tikzpicture}
    \begin{axis}[
        width=\textwidth, height=6cm,     
        grid = major,
	ybar,
        grid style={dashed, gray!30},
        xmin=0,     
        xmax=1,    
        ymin=0,     
        ymax=0.3,   
        /pgfplots/xtick={0,.2,...,1}, 
	/pgfplots/ytick={0,.1,...,.4},
	bar width=2pt,
        axis background/.style={fill=white},
        ylabel=Degree distribution,
        xlabel=Normalized degree,
        tick align=outside,
	legend style={
		cells={anchor=west},
		legend pos=north east,
		}
	]
 
      \addplot [color=red,fill=red] table {Learning_Convolutional_Fig_Data_N_40_K_20_L_50_Const_Deg_b.dat};

   \end{axis} 
\end{tikzpicture}
		\caption{Constraint neurons degrees\label{const_deg_80_40_60}}
    \end{subfigure}
	\caption{Pattern and constraint neuron degree distributions for $n=960$, $L=60$, and an average of $40$ constraints per cluster. The learning parameters are $\alpha_t \propto 0.95/t$, $\eta = 0.75/\alpha_t$ and $\theta_t \propto 0.05/t$.\label{sparsity_inside_cluster_N_80_K_40_L_60}}
\end{figure*}

Figure~\ref{sparsity_inside_cluster_N_80_K_40_L_60} illustrates the same results for a network of size $n=960$, which is divided into $60$ clusters, each with size $80$, on average. The learning parameters are the same as before, i.e., $\alpha_t \propto 0.95/t$, $\eta = 0.75/\alpha_t$ and $\theta_t \propto 0.05/t$, and each pattern neuron is connected to $5$ clusters on average. Note that the overall normalized degrees are smaller compared to the case of $n=400$, which indicates sparser clusters on average. 
In almost all cases that we have tried, the learning phase converges within two learning iterations, i.e., by going over the data set only twice.

\subsubsection{Recall Results} 
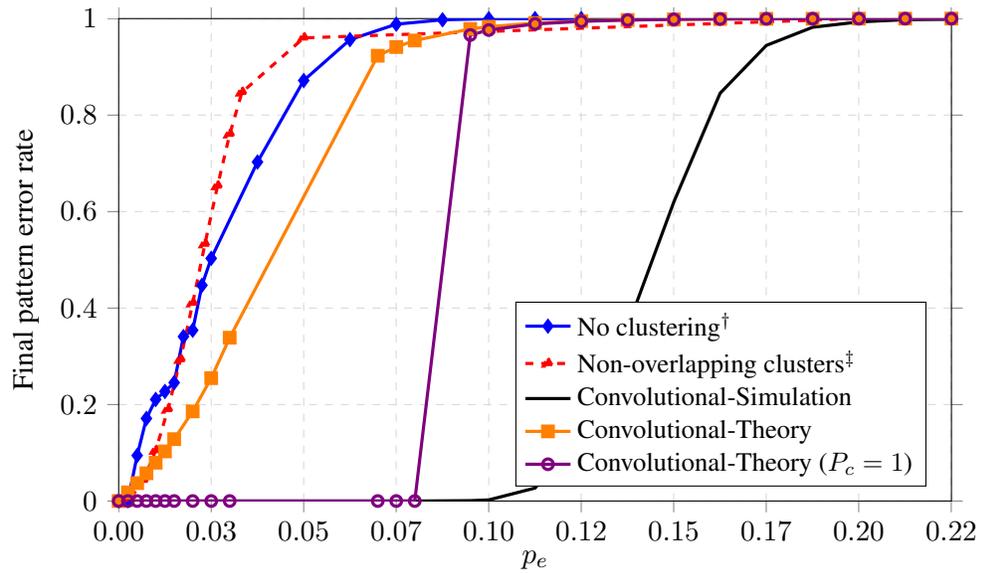
\begin{figure}
\centering
\centering
\begin{tikzpicture}[spy using outlines={blue, circle, 
                          connect spies}]
    \begin{axis}[
        width=\textwidth, height=8cm,     
        grid = major,
        grid style={dashed, gray!30},
        xmin=0,     
        xmax=.225,    
        ymin=0,     
        ymax=1.0,   
        /pgfplots/xtick={0,0.025,...,.225}, 
	 x tick label style={
        /pgf/number format/.cd,
        fixed,
        fixed zerofill,
        precision=2,
        /tikz/.cd
    	 },
        axis background/.style={fill=white},
        ylabel=Final pattern error rate,
        xlabel=$p_e$,
        tick align=outside,
	legend style={
		cells={anchor=west},
		legend pos=south east,
		font = \small,
		}
	]
      \addplot [color=blue,very thick,mark=diamond*] table {PER_Subspace_N_400_alpha_095_beta_1_theta_031_WTA.dat};
      \addplot [color=red,dashed,very thick,mark=triangle*] table {PER_MultiLevel_Fig_Data_N_100_Kt_100_Level_2.dat};
      \addplot [color=black,very thick,mark=circle*] table {PER_Convolutional_Fig_Data_N_40_K_20_L_50_Simulations.dat};
      \addplot [color=orange,very thick,mark=square*] table {PER_Convolutional_Fig_Data_N_40_K_20_L_50_Theory_New_P_c_real.dat};
      \addplot [color=violet,very thick,mark=o] table {PER_Convolutional_Fig_Data_N_40_K_20_L_50_Theory_New_P_c_1.dat};
      \legend{No clustering$^\dagger$ \\ Non-overlapping clusters$^\ddagger$\\ Convolutional-Simulation\\ Convolutional-Theory\\ Convolutional-Theory ($P_c = 1$)\\}
   \end{axis} 
\vspace{1cm}
\node at (2,-1.5) [rectangle]{$^\dagger$: \citep{KSS},};
\node at (6.3,-1.5) [rectangle]{$^\ddagger$: \citep{SK_ISIT2012}};

\end{tikzpicture}
\caption{Recall error rate along with theoretical bounds for different architectures of network with $n=400$ pattern neurons and $L=50$ clusters. We compare the performance of our method with two other constructions where either no notion of cluster was considered \citep{KSS} or no overlaps between clusters was assumed \citep{SK_ISIT2012}. \label{fig:PER_Convolutional_n_400}}
\end{figure}

Figure~\ref{fig:PER_Convolutional_n_400} illustrates the performance of the recall algorithm. The horizontal and vertical axes represent the average fraction of erroneous neurons and the final Pattern Error Rate (PER), respectively. The performance is compared against the theoretical bound derived in Theorem~\ref{th:outside} as well as the the two constructions proposed by \citet{KSS} and \citet{SK_ISIT2012}. The parameters used for this simulation are $n = 400$, $L = 50$ and $\varphi = 0.82$. For the \emph{non-overlapping} clusters approach proposed by \citet{SK_ISIT2012}, the network size is $n = 400$ with $4$ clusters in the first level and one cluster in the second level (identical to their simulations). The convolutional neural network proposed in this paper clearly outperforms the prior art.
Note that for the theoretical estimates used in Figure \ref{fig:PER_Convolutional_n_400}, we both calculated the probability of correcting a single error by each cluster $P_c$ (via the lower bound in Theorem \ref{theorem_algo_recall_within}), and by fixing it to $P_c=1$. The corresponding curves in Figure~\ref{fig:PER_Convolutional_n_400} show that the later estimate is tighter, i.e., when each cluster can correct a single error with probability close to $1$.

\begin{figure}
\centering
\centering
\begin{tikzpicture}[spy using outlines={blue, circle, 
                          connect spies}]
    \begin{axis}[
        width=\textwidth, height=6cm,     
        grid = major,
        grid style={dashed, gray!30},
        xmin=0,     
        xmax=0.25,    
        ymin=0,     
        ymax=1.0,   
        /pgfplots/xtick={0,.05,...,.251}, 
        axis background/.style={fill=white},
        ylabel=Final pattern error rate,
        xlabel=$p_e$,
	x tick label style={
        /pgf/number format/.cd,
        fixed,
        fixed zerofill,
        precision=2,
        /tikz/.cd
    	 },
        tick align=outside,
	legend style={
		cells={anchor=west},
		legend pos=south east,
		}
	]
\addplot [color=blue,very thick,mark=triangle*] table {PER_Convolutional_Fig_Data_N_40_K_20_L_50_Simulations.dat};
      \addplot [color=blue,dashed,very thick,mark=diamond*] table {PER_Convolutional_Fig_Data_N_40_K_20_L_50_Theory_New_P_c_real.dat};
\addplot [color=red,very thick,mark=*] table {PER_Convolutional_Fig_Data_N_80_K_40_L_60_Simulations.dat};
      \addplot [color=red,dashed,very thick,mark=square*] table {PER_Convolutional_Fig_Data_N_80_K_40_L_60_Theory_New_P_c_real.dat};
      \legend{$n=400$-Simulation\\ $n=400$-Theory\\ $n=960$-Simulation\\ $n=960$-Theory\\}
   \end{axis}

\end{tikzpicture}
\caption{Recall error rate and the theoretical bounds for different architectures of network with $n=960$ and $n=400$ pattern neurons and $L=60$ and $L=50$ clusters, respectively..\label{fig:PER_Convolutional_n_960}}
\end{figure}
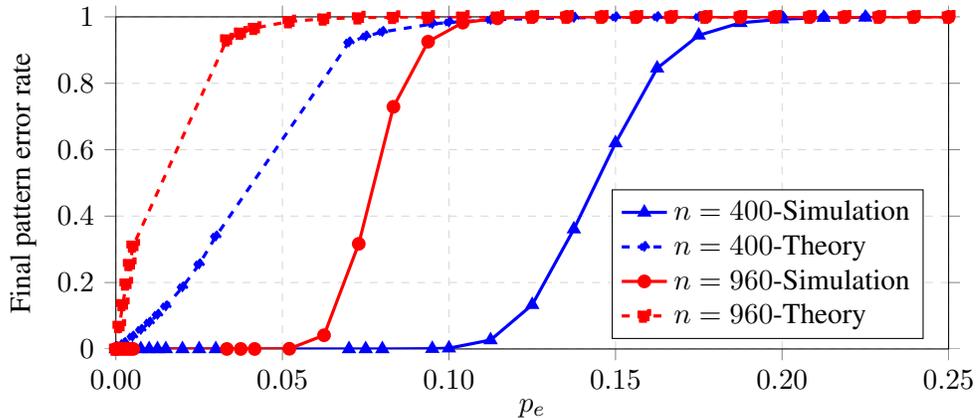

Figure~\ref{fig:PER_Convolutional_n_960} shows the final PER for the network with $n=960$ and $L=60$ clusters. Comparing the PER with that of a network with $n=400$ neurons and $L=50$ clusters, we witness a degraded performance. At first glance this might seem surprising as we increased both the network size and the number of clusters. However, the key point in determining the performance of Algorithm~\ref{algo:peeling} is not the number of clusters but rather the size of the clusters and the cluster nodes degree distribution $\widetilde{\rho}(x)$. In the network with $n=960$, we have around $80$ pattern neurons per cluster, while in the network with $n=400$ we have around $n=40$ neurons per cluster. Clearly, by increasing the network size without increasing the number of clusters, the chance of a cluster experiencing more than one error increases (remember, each cluster can correct a single error). This in turn results in an inferior performance. Hence, increasing the network size helps only if the number of clusters are increased correspondingly.
%


\subsection{Real Datasets}\label{sec:convolutional_simulation_real_dataset} 

So far, we have tested our proposed method over synthetic datasets where we generated patterns in such a way that they all belong to a subspace. In many real datasets (e.g., images and natural sounds), however, patterns rarely form a subspace. Rather, due to their common structures, they come very close to forming one. The focus of this section is to show how our proposed method can be adapted to such scenarios.

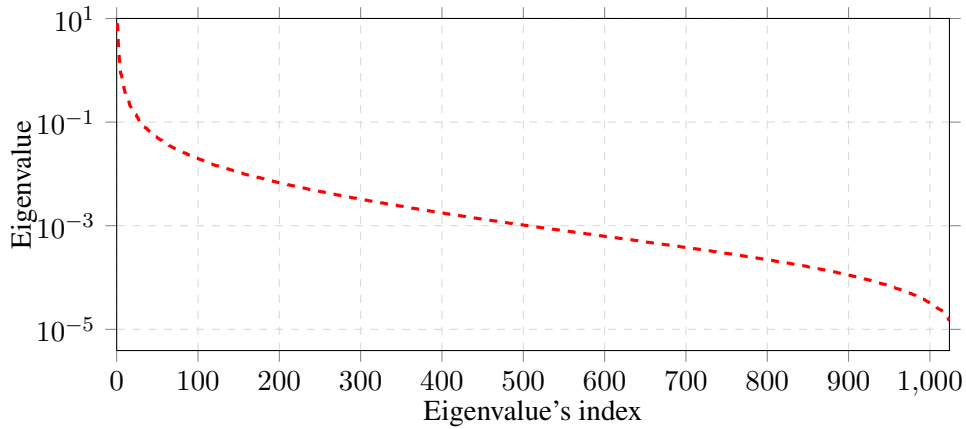
\begin{figure}
\begin{center}
\centering
\begin{tikzpicture}
    \begin{axis}[
        width=\textwidth, height=6cm,     
        grid = major,
        grid style={dashed, gray!30},
        ymode=log,log basis y=10,
        xmin=0,     
        xmax=1024,    
        ymin=0,     
        ymax=10,   
        /pgfplots/xtick={0,100,...,1024}, 
        axis background/.style={fill=white},
        xlabel=Eigenvalue's index,
        ylabel=Eigenvalue,
        tick align=outside,
	]
 
      \addplot [color=red,dashed,very thick] table {Eigenvalue_CIFAR_10_10000_Gray.dat};
   \end{axis} 
\end{tikzpicture}
\caption{The eigenvalues of a dataset with $1000$ gray-scale images of size $32\times 32$, uniformly sampled from $10$ classes of the CIFAR-10 dataset \citep{cifar10}.}
\label{fig:eigenvalues_cifar_10}
\end{center}
\end{figure}

More specifically, let $\mc{X}$ denote a dataset of $C$ patterns of length $n$. Here we assume that patterns are all vectorized and form the rows of the matrix $\mc{X}$. The eigenvalues of the correlation matrix $A = \mc{X}^\top \mc{X}$ indicate how close the patterns are to a subspace. Note that $A$ is a positive semidefinite matrix, so all eigenvalues are non-negative. In particular, if we have an eigenvalue 0 with positive multiplicity, then the patterns belong to a subspace. Similarly, if we have a set of eigenvalues all close to zero, then the patterns are close to a subspace of the $n$-dimensional space. 
%
 Figure \ref{fig:eigenvalues_cifar_10} illustrates the eigenvalue distribution of the correlation matrix for a dataset of $C=10000$ gray-scale images of size $32 \times 32$, sampled from $10$ classes of the CIFAR-10 dataset \citep{cifar10}. Each image is quantized to $16$ levels. Based on our notation, $n=1024$ and $Q=16$. As evident from the figure, almost half of the $1024$ eigenvalues are less than $0.001$, suggesting that the patterns are very close to a subspace.

\subsubsection{Simulation Scenario}

In order to adapt our method to this new scenario where patterns \emph{approximately} belong to a subspace, we need to slightly modify the learning and the recall algorithms. We use CIFAR-10 datase as the running example, however, the principles described below can be easily applied to other datasets. 

To start, we first alter the way patterns are represented in such a way that is makes them easier to learn for our algorithm. More specifically, since the images are quantized to $16$ levels, we can represent a $16$-level pixel with $4$ bits. As such, instead of having $1024$ integer-valued pattern neurons to represent the patterns in the dataset, we will have $4096$ binary pattern neurons. We adopt this modified description as it facilitates the learning process. 
%
%

We then apply Algorithm~\ref{algo_learning} as before to learn the patterns in the dataset. Obviously, since the patterns do not exactly form a subspace, we cannot expect the algorithm to finish with a weight vector $w$ that is \emph{orthogonal} to all the patterns. Nevertheless, by applying the learning algorithm, we will have a weight vector whose projection on the patterns is rather small. Following the same procedure as before, we obtain $L$ neural graphs, $W^{(1)}, \dots,W^{(L)}$, for each of the $L$ clusters. Note again that $W^{(i)}$'s are approximately (rather than exactly) orthogonal to the (sub-)patterns. 

\begin{figure}[t!]
\begin{center}
\begin{minipage}{0.75\textwidth}
\begin{subfigure}{0.32\textwidth}
        \centering
        \includegraphics[width=.95\textwidth]{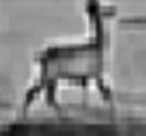}
    \end{subfigure}~
\begin{subfigure}{0.32\textwidth}
        \centering
        \includegraphics[width=.95\textwidth]{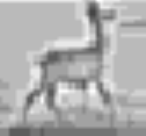}
    \end{subfigure}~
\begin{subfigure}{0.31\textwidth}
        \centering
        \includegraphics[width=.94\textwidth]{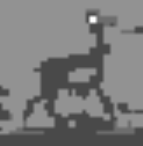}
    \end{subfigure}~
\end{minipage}
\\
\begin{minipage}{0.75\textwidth}
\setcounter{subfigure}{0}
\begin{subfigure}{0.32\textwidth}
        \centering
        \includegraphics[width=.95\textwidth]{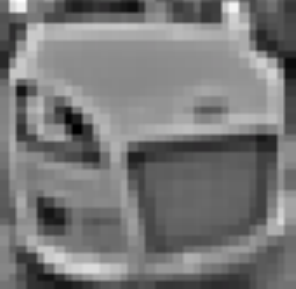}
        \caption{Original image}
    \end{subfigure}~
\begin{subfigure}{0.32\textwidth}
        \centering
        \includegraphics[width=.95\textwidth]{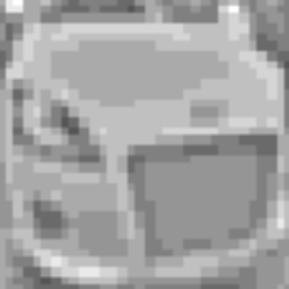}
        \caption{Quantized image}
    \end{subfigure}~
\begin{subfigure}{0.32\textwidth}
        \centering
        \includegraphics[width=.95\textwidth]{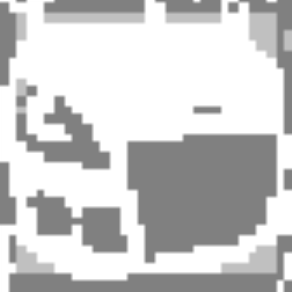}
        \caption{Learned image}
    \end{subfigure}~
\end{minipage}
\caption{Original vs. learned images}
\label{fig:learned_images}
\end{center}
\end{figure}

Our main observation is the following. We can interpret the deviation from the subspace as noise. Consequently, if we apply the intra-cluster recall method (Algorithm \ref{algo:correction}) to the patterns, we can find out what the network has actually learned in response to original patterns from the dataset $\mc{X}$. In other words, Algorithm \ref{algo:correction} identifies the projection of original patterns to a subspace $\mc{X}^\prime$. Hence, all the learned patterns are orthogonal to the connectivity matrix $W$. This idea is shown in figure~\ref{fig:learned_images}, where we have the original image (left), the quantized version (middle), and the image learned by the proposed algorithm. It is worth observing that what the network has learned focuses more on the actual objects rather than unnecessary details.


%
%
%

%

For the recall phase, the approach is similar to before: we are given a set of noisy patterns and the goal is to retrieve the correct versions. In our simulations, we assume that the noise is added to the learned patterns (see above) but we can also consider the situation where noise is added to the quantized patterns. 

%
%
%
%
%

\subsubsection{Learning Results}\label{sec:learning-real}
\begin{figure}
\centering
\centering
\begin{tikzpicture}[spy using outlines={blue, circle, 
                          connect spies}]
    \begin{axis}[
  	 scaled ticks=false,
        width=\textwidth, height=6cm,     
        grid = major,
        grid style={dashed, gray!30},
        xmin=0,     
        xmax=27,    
        ymin=0,     
        ymax=0.006,   
        /pgfplots/xtick={0,3,...,27}, 
	 y tick label style={
	at={(-10,0.5)},
        /pgf/number format/.cd,
        fixed,
        fixed zerofill,
        precision=3,
        /tikz/.cd
    	 },
        axis background/.style={fill=white},
        ylabel= Learning cost,
        xlabel= $t$,
	legend style={
		cells={anchor=west},
		legend pos=south east,
		font = \small,
		}
	]
     \addplot [color=red,very thick,mark=diamond*] table {Learn_Cost_Fig_Data_CIFAR_10000_n_100_L_60_Simul.dat};

   \end{axis} 
\end{tikzpicture}
\caption{Average cost versus time for learning a weight vector in a network with $n=4096$ pattern neurons and $L=401$ clusters. The size of clusters is set to $100$ with (around) $50$ constraints in each cluster. \label{fig:learning_cost_CIFAR_10000}}
\end{figure}
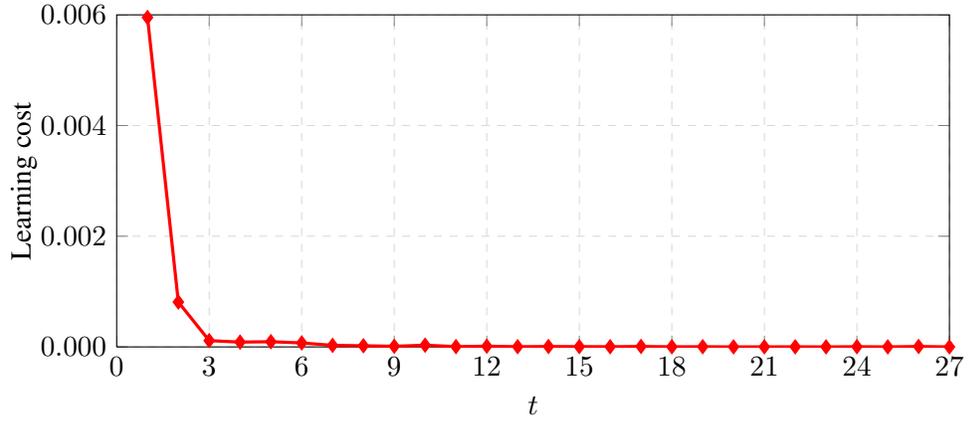

Figure \ref{fig:learning_cost_CIFAR_10000} illustrates the average cost (defined by $E(w^{(\ell)}(t))$ in Section \ref{sec:convergence_learning}) for learning one constraint vector versus the number of iterations. In this example, the learning parameters are $\alpha_0 = 0.95/t$, $\eta = 1$ and $\theta_t = 0.01/t$. The considered neural netowrk has $n=4096$ pattern neurons and $L =401$ clusters of size $100$. The learning process terminates if a) an orthogonal weight vector is found or b) $200$ iterations is done.

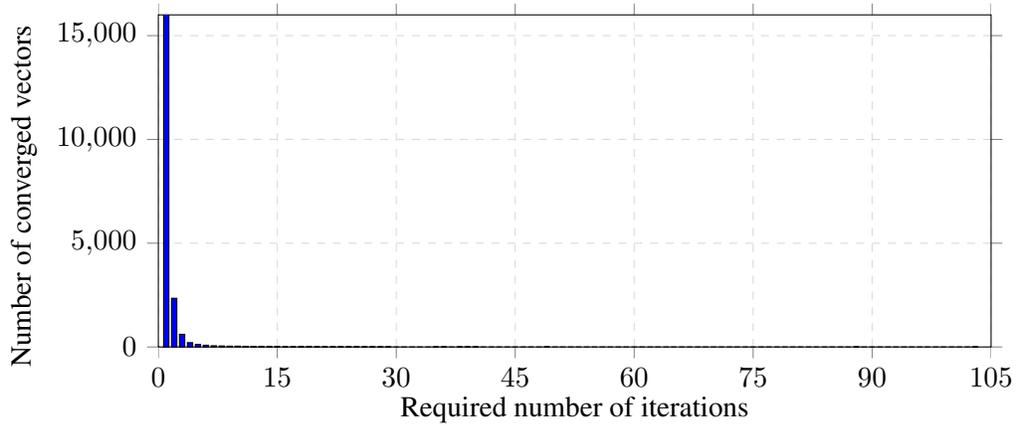
\begin{figure}
\centering
\centering
\begin{tikzpicture}
    \begin{axis}[
  	 scaled ticks=false,
        width=\textwidth, height=6cm,     
        grid = major,
	 ybar,
        grid style={dashed, gray!30},
        xmin=0,     
        xmax=105,    
        ymin=0,     
        ymax=16000,   
        /pgfplots/xtick={0,15,...,105}, 
	/pgfplots/ytick={0,5000,10000,15000},
	bar width=2pt,
        axis background/.style={fill=white},
	y label style={at={(axis description cs:-0.05,1)},rotate=0,anchor=east},
        ylabel=Number of converged vectors,
        xlabel=Required number of iterations,
        tick align=outside,
	legend style={
		cells={anchor=west},
		legend pos=north east,
		}
	]
      \addplot [color=black,fill=blue] table {Learn_Itr_Fig_Data_CIFAR_10000_n_100_L_60_Simul.dat};
   \end{axis} 
\end{tikzpicture}
\caption{Number of the iterations required for Algortihm \ref{algo_learning} to learn a vector orthogonal to the patterns in the dataset $\mc{X}^\prime$, in a network with $n=4096$ pattern neurons and $L=401$ clusters of size $100$ and (around) $50$ constraints in each cluster.\label{fig:learning_itr_CIFAR_10000}}
\end{figure}

Figure\ref{fig:learning_itr_CIFAR_10000} illustrates the required number of iterations of Algorithm \ref{algo_learning} so that a weight vector orthogonal the patterns in the dataset $\mc{X}^\prime$ is obtained. As we see from the figure, in the majority of the cases, one pass over the dataset is enough. As before, we have $\alpha_0 = 0.95/t$, $\eta = 1$ and $\theta_t = 0.01/t$.

Furthermore, we have also uploaded a short video clip of the learning algorithm in action (i.e., iteration by iteration) for a few sample images from the dataset. The clip is available through the following link: \url{http://goo.gl/evcNOh}.

\subsubsection{Recall Results} 
\begin{figure}
\centering
\centering
\begin{tikzpicture}[spy using outlines={blue, circle, 
                          connect spies}]
    \begin{axis}[
        width=\textwidth, height=6cm,     
        grid = major,
        grid style={dashed, gray!30},
        xmin=0,     
        xmax=.008,    
        ymin=0,     
        ymax=1.0,   
        /pgfplots/xtick={0,0.002,...,.008}, 
	 x tick label style={
        /pgf/number format/.cd,
        fixed,
        fixed zerofill,
        precision=3,
        /tikz/.cd
    	 },
        axis background/.style={fill=white},
        ylabel=Final pattern error rate,
        xlabel=$p_e$,
        tick align=outside,
	legend style={
		cells={anchor=west},
		legend pos=south east,
		font = \small,
		}
	]
      \addplot [color=orange,very thick,mark=diamond*] table {PER_Convolutional_Fig_Data_CIFAR_5000_n_100_L_501_Simul.dat};
      \addplot [color=blue,very thick,mark=diamond*] table {PER_Convolutional_Fig_Data_CIFAR_10000_n_100_L_60_Simul.dat};      
      \addplot [color=red,very thick,mark=diamond*] table {PER_Convolutional_Fig_Data_CIFAR_50000_n_90_L_502_Simul.dat};      
      \legend{$C=5000$\\$C=10000$\\$C=50000$\\}
   \end{axis} 
\end{tikzpicture}
\caption{Recall error rate for a network with $n=4096$ pattern neurons and $L=401$ clusters, with cluster size equal to $n_\ell = 100$. The proposed recal method was applied to a dataset of $10000$ images sampled from the CIFAR-10 database\label{fig:PER_Convolutional_CIFAR_10000}}
\end{figure}
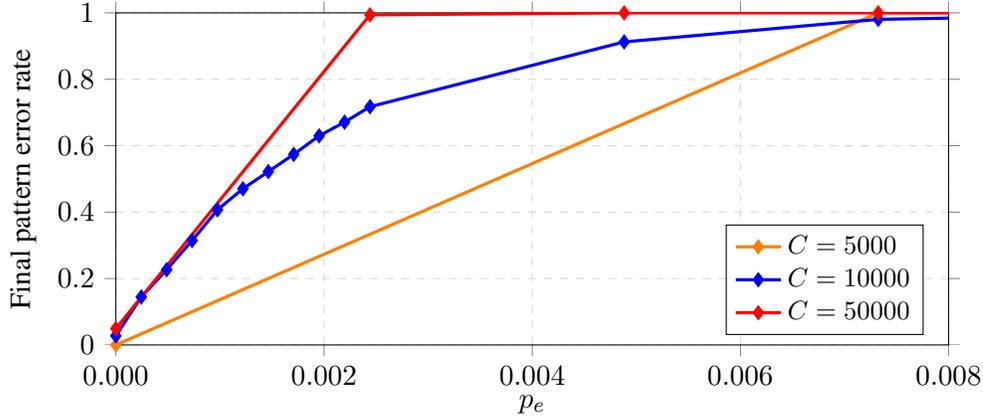

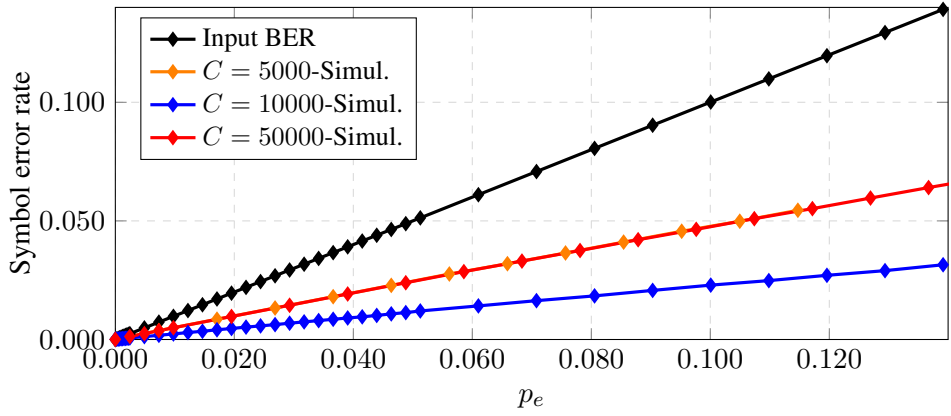
\begin{figure}
\centering
\centering
\begin{tikzpicture}[spy using outlines={blue, circle, 
                          connect spies}]
    \begin{axis}[
        width=\textwidth, height=6cm,     
        grid = major,
        grid style={dashed, gray!30},
        xmin=0,     
        xmax=.14,    
        ymin=0,     
        ymax=0.14,   
        /pgfplots/xtick={0,0.02,...,.139}, 
	 x tick label style={
        /pgf/number format/.cd,
        fixed,
        fixed zerofill,
        precision=3,
        /tikz/.cd
    	 },
	y tick label style={
	at={(-10,0.5)},
        /pgf/number format/.cd,
        fixed,
        fixed zerofill,
        precision=3,
        /tikz/.cd,
    	 },
        axis background/.style={fill=white},
        ylabel=Symbol error rate,
        xlabel=$p_e$,
	legend style={
		cells={anchor=west},
		legend pos=north west,
		font = \small,
		}
	]
     \addplot [color=black,very thick,mark=diamond*] table {BER_Convolutional_Fig_Data_CIFAR_10000_n_100_L_60_Input_BER.dat};
      \addplot [color=orange,very thick,mark=diamond*] table {BER_Convolutional_Fig_Data_CIFAR_5000_n_100_L_501_Simul.dat};     
      \addplot [color=blue,very thick,mark=diamond*] table {BER_Convolutional_Fig_Data_CIFAR_10000_n_100_L_60_Simul.dat};
      \addplot [color=red,very thick,mark=diamond*] table {BER_Convolutional_Fig_Data_CIFAR_50000_n_90_L_502_Simul.dat};            
      \legend{Input BER\\$C=5000$-Simul.\\$C=10000$-Simul.\\$C=50000$-Simul.\\}
   \end{axis} 
\end{tikzpicture}
\caption{Symbol error rates for a network with $n=4096$ pattern neurons and $L=401$ clusters, applied to a dataset of $10000$ images sampled from the CIFAR-10 database.\label{fig:BER_Convolutional_CIFAR_10000}}
\end{figure}

Figures~\ref{fig:PER_Convolutional_CIFAR_10000} and \ref{fig:BER_Convolutional_CIFAR_10000} show the recall error rate as we increase the noise level for the neural network used in Sec~\ref{sec:learning-real}. The update thresholds for the recall algorithm are set to $\varphi = 0.85$ and $\psi = 0.005$. The inter-module recall procedure (Algorithm~\ref{algo:peeling}) is performed at most $80$ times and the corresponding error rates are calculated by evaluating the difference between the final state of pattern neurons (after running Algorithm~\ref{algo:peeling}) and the noise-free patterns in the dataset $\mc{X}^\prime$.

%

\begin{figure}[t!]
\begin{center}
\begin{minipage}{0.85\textwidth}
\begin{subfigure}{0.2\textwidth}
        \centering
        \includegraphics[width=.95\textwidth]{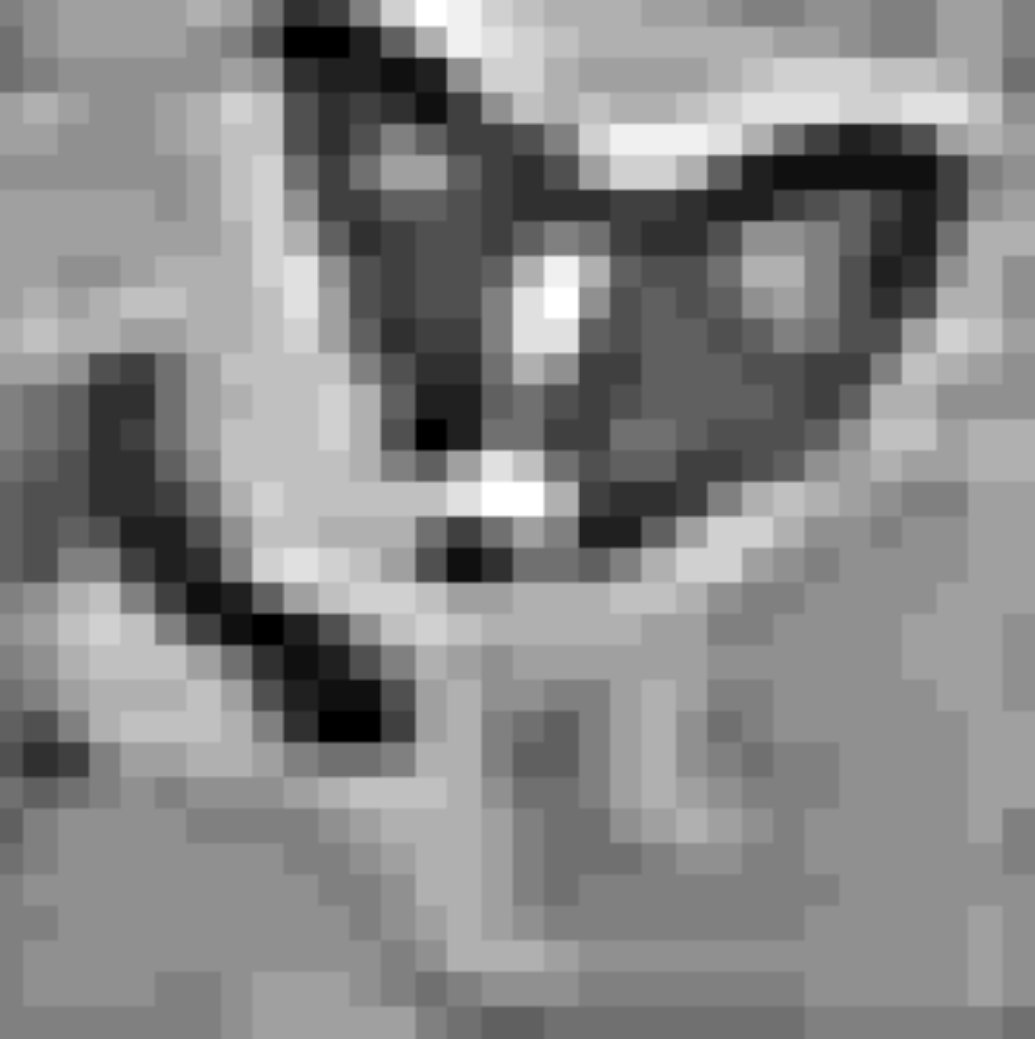}
\caption{Original image}
    \end{subfigure}~
\begin{subfigure}{0.2\textwidth}
        \centering
        \includegraphics[width=.95\textwidth]{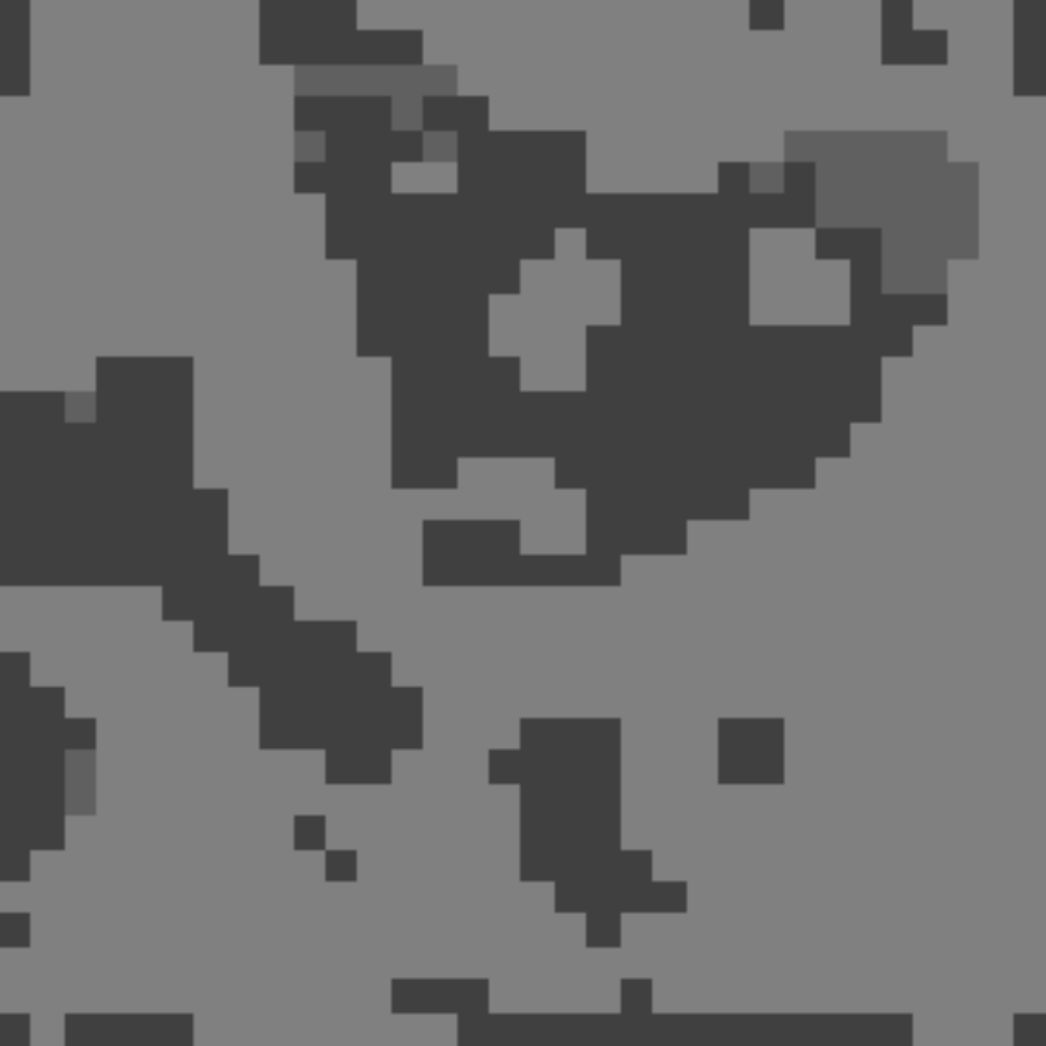}
\caption{Learned image}
    \end{subfigure}~
\begin{subfigure}{0.2\textwidth}
        \centering
        \includegraphics[width=.95\textwidth]{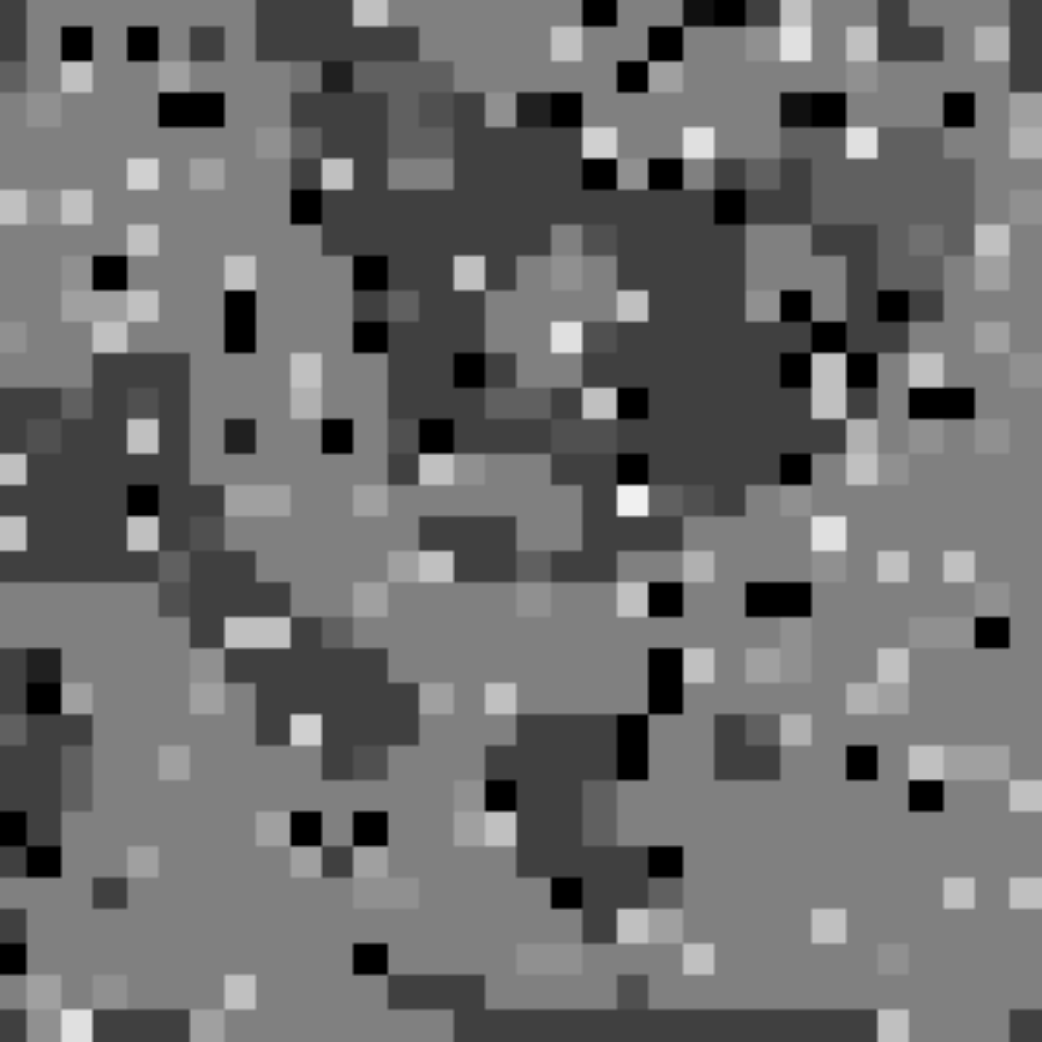}
\caption{$\hbox{SNR}_{\hbox{I}}: 2.10$}
    \end{subfigure}~
\begin{subfigure}{0.2\textwidth}
        \centering
        \includegraphics[width=.95\textwidth]{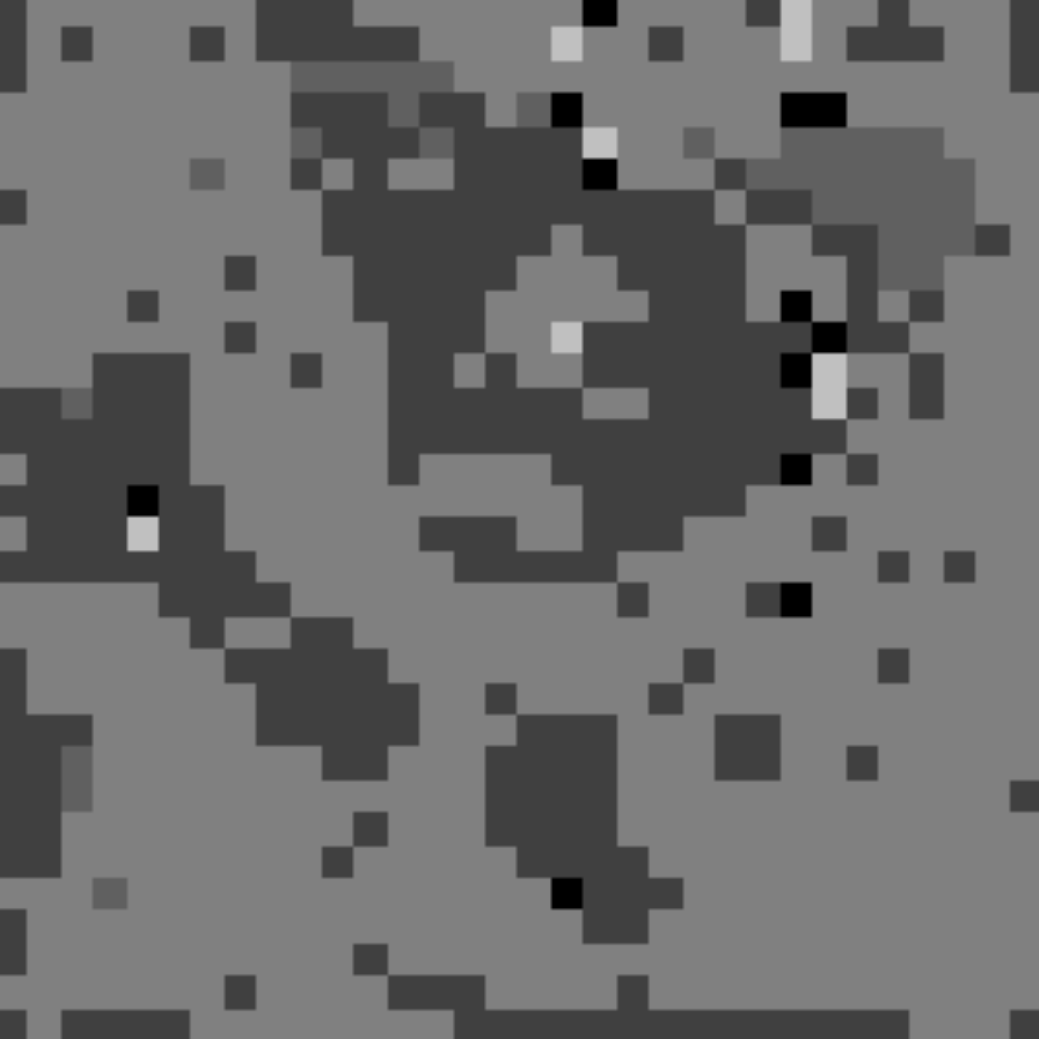}
\caption{$\hbox{SNR}_{\hbox{O}}: 2.84$}
    \end{subfigure}
\\
\vspace{1cm}
\hspace{-.25cm}
\setcounter{subfigure}{0}
\begin{subfigure}{0.2\textwidth}
        \centering
        \includegraphics[width=.95\textwidth]{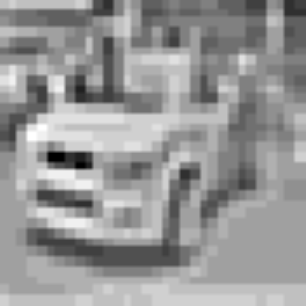}
\caption{Original image}
    \end{subfigure}~
\begin{subfigure}{0.2\textwidth}
        \centering
        \includegraphics[width=.95\textwidth]{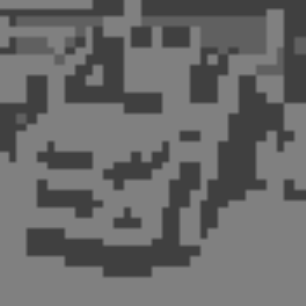}
\caption{Learned image}
    \end{subfigure}~
\begin{subfigure}{0.2\textwidth}
        \centering
        \includegraphics[width=.95\textwidth]{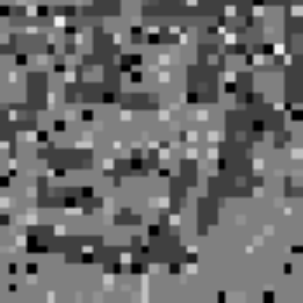}
\caption{$\hbox{SNR}_{\hbox{I}}: 2.15$}
    \end{subfigure}~
\begin{subfigure}{0.2\textwidth}
        \centering
        \includegraphics[width=.95\textwidth]{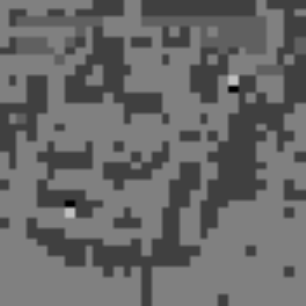}
\caption{$\hbox{SNR}_{\hbox{O}}: 2.93$}
    \end{subfigure}
\\
\vspace{.5cm}
\hspace{-.25cm}
\setcounter{subfigure}{0}
\begin{subfigure}{0.2\textwidth}
        \centering
        \includegraphics[width=.95\textwidth]{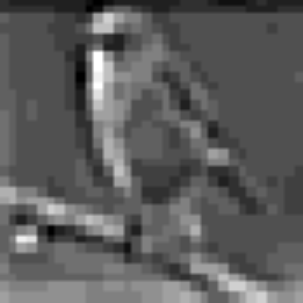}
\caption{Original image}
    \end{subfigure}~
\begin{subfigure}{0.2\textwidth}
        \centering
        \includegraphics[width=.95\textwidth]{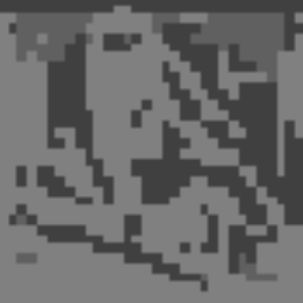}
\caption{Learned image}
    \end{subfigure}~
\begin{subfigure}{0.2\textwidth}
        \centering
        \includegraphics[width=.95\textwidth]{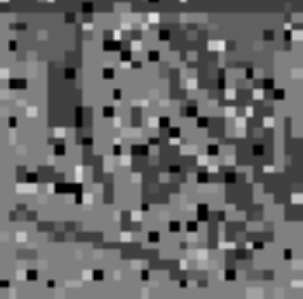}
\caption{$\hbox{SNR}_{\hbox{I}}: 2.07$}
    \end{subfigure}~
\begin{subfigure}{0.2\textwidth}
        \centering
        \includegraphics[width=.95\textwidth]{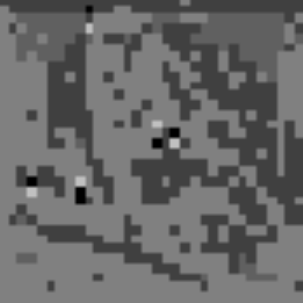}
\caption{$\hbox{SNR}_{\hbox{O}}: 3.16$}
    \end{subfigure}
\\
\vspace{.5cm}
\hspace{-.25cm}
\setcounter{subfigure}{0}
\begin{subfigure}{0.2\textwidth}
        \centering
        \includegraphics[width=.95\textwidth]{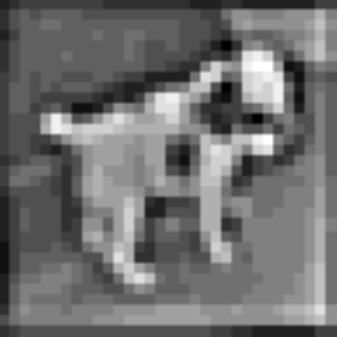}
\caption{Original image}
    \end{subfigure}~
\begin{subfigure}{0.2\textwidth}
        \centering
        \includegraphics[width=.95\textwidth]{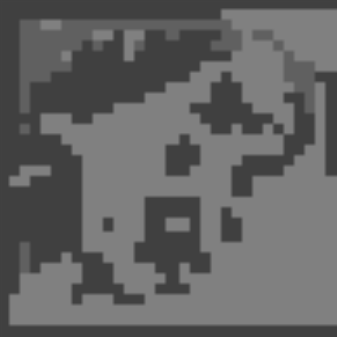}
\caption{Learned image}
    \end{subfigure}~
\begin{subfigure}{0.2\textwidth}
        \centering
        \includegraphics[width=.95\textwidth]{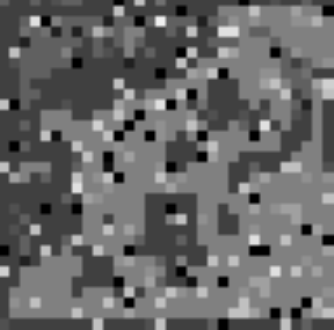}
\caption{$\hbox{SNR}_{\hbox{I}}: 2.09$}
    \end{subfigure}~
\begin{subfigure}{0.2\textwidth}
        \centering
        \includegraphics[width=.95\textwidth]{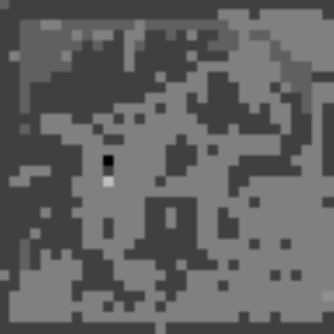}
\caption{$\hbox{SNR}_{\hbox{O}}: 2.96$}
    \end{subfigure}
\\
\vspace{.5cm}
\hspace{-.25cm}
\setcounter{subfigure}{0}
\begin{subfigure}{0.2\textwidth}
        \centering
        \includegraphics[width=.95\textwidth]{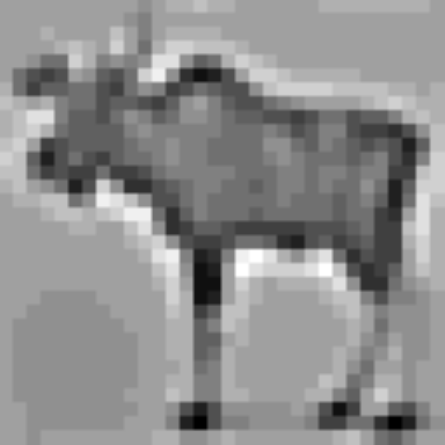}
\caption{Original image}
    \end{subfigure}~
\begin{subfigure}{0.2\textwidth}
        \centering
        \includegraphics[width=.95\textwidth]{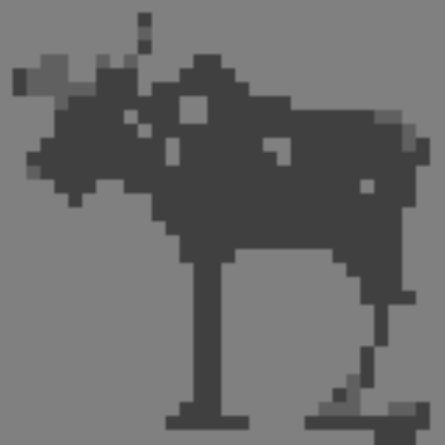}
\caption{Learned image}
    \end{subfigure}~
\begin{subfigure}{0.2\textwidth}
        \centering
        \includegraphics[width=.95\textwidth]{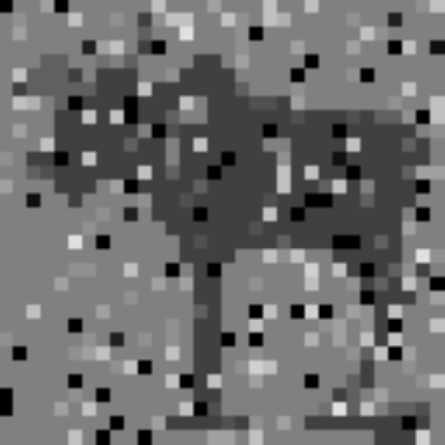}
\caption{$\hbox{SNR}_{\hbox{I}}: 2.08$}
    \end{subfigure}~
\begin{subfigure}{0.2\textwidth}
        \centering
        \includegraphics[width=.95\textwidth]{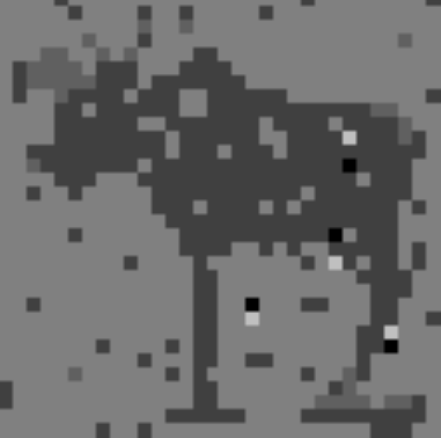}
\caption{$\hbox{SNR}_{\hbox{O}}: 2.81$}
    \end{subfigure}~
\end{minipage}
\caption{Examples of the learning and recall phase for images sampled from CIFAR-10. Here, $\hbox{SNR}_{\hbox{I}}$ and $\hbox{SNR}_{\hbox{O}}$ denote the input and output SNR's, respectively. }
\label{fig:recall_results_examples_CIFAR_10_10000}
\end{center}
\end{figure}

Figure \ref{fig:recall_results_examples_CIFAR_10_10000} illustrates a few instances of the recalled images. In this figure, we have the original images (first column), the learned images (second column), the noisy versions (third column), and the recalled images (forth column). The figure also shows the input and the output Signal to Noise Ratios (SNR) for each example. Note that in all examples the SNR increases as we apply our recall algorithm. For this example we chose $\varphi = 0.95$ and $\psi = 0.025$. We have also uploaded a short video clip of the recall algorithm in action for a few sample images from the dataset, which can be found on \url{http://goo.gl/EHJfds}.



\section{Analysis} \label{sec:analysis}
This section contains the proofs of all theorems and technical lemmas we used in the paper.
\subsection{Proof of Lemma~\ref{lem:avoiding_all_zero}}

We proceed by induction. To this end, assume $\Vert w^{(\ell)}(t) \Vert_2 >0$ and let $$\grave{w}^{(\ell)}(t) = w^{(\ell)}(t) - \alpha_t y^{(\ell)}(t)\left(x^{(\ell)}(t) - \frac{y^{(\ell)}(t) w^{(\ell)}(t)}{\Vert w^{(\ell)}(t) \Vert_2^2}\right).$$ Note that $$\Vert \grave{w}^{(\ell)}(t) \Vert_2^2 = \Vert w^{(\ell)}(t) \Vert_2^2 + \alpha_t^2 y^{(\ell)}(t)^2 \Vert x^{(\ell)}(t) - \frac{y^{(\ell)}(t) w^{(\ell)}(t)}{\Vert w^{(\ell)}(t) \Vert_2^2} \Vert_2^2 \geq \Vert w^{(\ell)}(t) \Vert_2^2 > 0.$$ Now,
\beqa
\Vert w^{(\ell)}(t+1)\Vert_2^2 &=& \Vert \grave{w}^{(\ell)}(t) \Vert^2+ \alpha_t^2 \eta^2 \Vert \Gamma(w^{(\ell)}(t),\theta_t) \Vert^2 - 2\alpha_t \eta \langle \Gamma(w^{(\ell)}(t),\theta_t), \grave{w}^{(\ell)}(t) \rangle  \nonumber \\
&\geq & \Vert \grave{w}^{(\ell)}(t) \Vert_2^2 + \alpha_t^2 \eta^2 \Vert \Gamma(w^{(\ell)}(t),\theta_t) \Vert^2 - 2\alpha_t \eta \Vert \Gamma(w^{(\ell)}(t),\theta_t)\Vert_2 \Vert \grave{w}^{(\ell)}(t)\Vert_2 \nonumber \\
&=& \left( \Vert \grave{w}^{(\ell)}(t) \Vert_2 - \alpha_t \eta \Vert \Gamma(w^{(\ell)}(t),\theta_t)\Vert_2 \right)^2 \nonumber
\eeqa
Thus, in order to have $\Vert w^{(\ell)}(t+1)\Vert_2 > 0$, we must have that $$ \Vert \grave{w}^{(\ell)}(t) \Vert_2 - \alpha_t \eta \Vert \Gamma(w^{(\ell)}(t),\theta_t)\Vert_2 > 0.$$ Given that $$\Vert\Gamma(w^{(\ell)}(t),\theta_t)\Vert_2 \leq \Vert w(t) \Vert_2 \leq \Vert \grave{w}^{(\ell)}(t) \Vert_2,$$ it is sufficient to have $ \alpha_t \eta < 1$ in order to achieve the desired inequality. This proves the lemma.

\subsection{Proof of Theorem~\ref{lemma_convergence}}
Let us define the \textit{correlation} matrix for the sub-patterns that lie within the domain of cluster $\ell$ as follows
$$A^{(\ell)} := \mb{E}\{x^{(\ell)} (x^{(\ell)})^T|x\in \mc{X}\}.$$ Also, let us define $$A^{(\ell)}_t := x^{(\ell)}(t)(x^{(\ell)}(t))^\top.$$ Hence, we have $A^{(\ell)} = \mb{E}(A^{(\ell)}_t)$. Furthermore, recall the learning cost function
$$E(t) = E(w^{(\ell)}(t)) =\frac{1}{C} \sum_{\mu =1}^{C} \left( \langle w^{(\ell)}(t), x^\mu \rangle \right)^2.$$
 From Alg.~\ref{algo_learning} we have
\bed
w^{(\ell)}(t+1) = w^{(\ell)}(t) - \alpha_t \left(y^{(\ell)}(t) \left(x^{(\ell)}(t) - \frac{y^{(\ell)}(t) w^{(\ell)}(t)}{\Vert w^{(\ell)}(t)\Vert_2^2}\right) + \eta \Gamma(w^{(\ell)}(t),\theta_t) \right).
\eed
Let $$Y^{(\ell)}(t) = \mb{E}_x (\mc{X}^{(\ell)} w^{(\ell)}(t)),$$ where $\mb{E}_x(\cdot)$ is the expectation over the choice of pattern $x(t)$ and $\mc{X}^{(\ell)}$ is the matrix of all the sub-patterns corresponding to cluster $\ell$ in the dataset. Thus, we will have
\beqa
Y^{(\ell)}(t+1) &=& Y^{(\ell)}(t) \left(1+ \alpha_t \frac{ \left(w^{(\ell)}(t)\right)^\top A^{(\ell)} w^{(\ell)}(t)}{\Vert w^{(\ell)}(t)\Vert_2^2}\right) -\alpha_t \left( \mc{X}^{(\ell)} A^{(\ell)} w^{(\ell)}(t) + \eta \mc{X}^{(\ell)} \Gamma(w^{(\ell)}(t),\theta_t) \right). \nonumber
\eeqa
Noting that $E(t) = \frac{1}{C} \Vert Y^{(\ell)}(t) \Vert_2^2$, we obtain
\beqa
E(t+1) &=&E(t)\left(1+ \alpha_t \frac{\left(w^{(\ell)}(t)\right)^\top A^{(\ell)} w^{(\ell)}(t)}{\Vert w^{(\ell)}(t)\Vert_2^2}\right)^2 \nonumber \\
&+& \frac{\alpha_t^2}{C} \Vert \mc{X}^{(\ell)} A^{(\ell)} w^{(\ell)}(t) + \eta \mc{X}^{(\ell)} \Gamma(w^{(\ell)}(t),\theta_t)\Vert_2^2 \nonumber \\
&-& 2\alpha_t \left(1+ \alpha_t \frac{\left(w^{(\ell)}(t)\right)^\top A^{(\ell)} w(t)}{\Vert w^{(\ell)}(t)\Vert_2^2}\right) \left(\left(w^{(\ell)}(t)\right)^\top \left(A^{(\ell)}\right)^2 w^{(\ell)}(t)\right) \nonumber \\
 &-& 2\alpha_t \left(1+ \alpha_t \frac{\left(w^{(\ell)}(t)\right)^\top A^{(\ell)} w^{(\ell)}(t)}{\Vert w^{(\ell)}(t)\Vert_2^2}\right) \left( \eta \left(w^{(\ell)}(t)\right)^\top A^{(\ell)} \Gamma(w^{(\ell)}(t),\theta_t)\right)
 \nonumber 
 \eeqa
 By omitting all the second order terms $O(\alpha_t^2)$, we obtain
\beqa 
E(t+1) &{\simeq}& E(t)\left(1+ 2\alpha_t \frac{\left(w^{(\ell)}(t)\right)^\top A^{(\ell)} w^{(\ell)}(t)}{\Vert w^{(\ell)}(t)\Vert_2^2}\right) \nonumber \\
&-& 2\alpha_t \left( \left(w^{(\ell)}(t)\right)^\top \left(A^{(\ell)}\right)^2 w^{(\ell)}(t) + \eta \left(w^{(\ell)}(t)\right)^\top A^{(\ell)} \Gamma(w^{(\ell)}(t),\theta_t)\right) \nonumber \\
&=&E(t)-2\alpha_t\left( \left(w^{(\ell)}(t)\right)^\top \left(A^{(\ell)}\right)^2 w^{(\ell)}(t)-\frac{ \left(w^{(\ell)}(t)\right)^\top A^{(\ell)} w^{(\ell)}(t)}{\Vert w^{(\ell)}(t)\Vert_2^2} E(t)\right)\nonumber \\
&-& 2\alpha_t \eta \left(w^{(\ell)}(t)\right)^\top A^{(\ell)} \Gamma(w^{(\ell)}(t),\theta_t) \label{eq:E(t+1)}
\eeqa
Note that 
\beqa
\alpha_t \eta \Vert \left(w^{(\ell)}(t)\right)^\top A^{(\ell)} \Gamma(w^{(\ell)}(t),\theta_t) \Vert_2 &\leq& \alpha_t \eta \Vert w^{(\ell)}(t) \Vert_2 \Vert A^{(\ell)} \Vert_2 \Vert \Gamma(w^{(\ell)}(t),\theta_t) \Vert_2 \nonumber \\
&\leq& \alpha_t \eta \Vert w^{(\ell)}(t) \Vert_2 \Vert A^{(\ell)} \Vert_2 (\sqrt{n}\theta_t).\nonumber
\eeqa
Since we have 
$\theta_t = \Theta(\alpha_t)$ and that $$\alpha_t \eta \Vert \left(w^{(\ell)}(t)\right)^\top A^{(\ell)} \Gamma(w^{(\ell)}(t),\theta_t) \Vert_2 = O(\alpha_t^2)$$
we can further simplify \eqref{eq:E(t+1)} as follows
\beqa 
E(t+1) &{\simeq}& E(t)-2\alpha_t\left( \left(w^{(\ell)}(t)\right)^\top \left(A^{(\ell)}\right)^2 w(t)-\frac{ \left(w^{(\ell)}(t)\right)^\top A^{(\ell)} \left(w^{(\ell)}(t)\right)}{\Vert w^{(\ell)}(t)\Vert_2^2}E(t)\right). \nonumber 
\eeqa

%

Thus, in order to show that the algorithm converges, we need to show that
\bed
\left( \left(w^{(\ell)}(t)\right)^\top \left(A^{(\ell)}\right)^2 w^{(\ell)}(t)-\frac{\left(w^{(\ell)}(t)\right)^\top A^{(\ell)} w^{(\ell)}(t)}{\Vert w^{(\ell)}(t)\Vert_2^2}E(t)\right) \geq 0
\eed
which in turn implies $E(t+1) \leq E(t)$. By noting that $$E(t) = \left(w^{(\ell)}(t)\right)^\top A^{(\ell)} w^{(\ell)}(t),$$ we must show that $$\left(w^{(\ell)}\right)^\top \left(A^{(\ell)}\right)^2 w^{(\ell)} \geq \left(\left(w^{(\ell)}\right)^\top A^{(\ell)} w^{(\ell)}\right)^2/\Vert w^{(\ell)}\Vert_2^2.$$ The left hand side is $\Vert A^{(\ell)}w^{(\ell)} \Vert_2^2$. For the right hand side, we have
\bed
\frac{\Vert \left(w^{(\ell)}\right)^\top A^{(\ell)} w^{(\ell)}\Vert_2^2}{\Vert w^{(\ell)} \Vert_2^2} \leq \frac{\Vert w^{(\ell)} \Vert_2^2 \Vert A^{(\ell)}w^{(\ell)} \Vert_2^2}{\Vert w^{(\ell)} \Vert_2^2} = \Vert A^{(\ell)}w^{(\ell)} \Vert_2^2.
\eed
The above inequality shows that $E(t+1) \leq E(t)$, which readily implies that for sufficiently large number of iterations, the algorithm converges to a local minimum $\hat{w}^{(\ell)}$ where $E(\hat{w}^{(\ell)}) = 0$. From Lemma \ref{lem:avoiding_all_zero} we know that $\Vert \hat{w}^{(\ell)} \Vert_2 > 0$. Thus, the only solution for $E(\hat{w}^{(\ell)}) = \Vert \mc{X}^{(\ell)} \hat{w}^{(\ell)} \Vert_2^2 =0$ is for $\hat{w}^{(\ell)}$ to be orthogonal to the patterns in the data set. 

\subsection{Proof of Theorem~\ref{theorem_algo_recall_within}}

In the case of a single error, we can easily show  that the noisy pattern neuron will always be updated towards the correct direction in Algorithm \ref{algo:correction}. For simplicity, let's assume the first pattern neuron of cluster $\ell$ is the noisy one. Furthermore, let $z^{(\ell)} = [1,0,\dots,0]$ be the noise vector. Denoting the $i^{th}$ column of the weight matrix by $W^{(\ell)}_i$, we will have $$y^{(\ell)} = \hbox{sign}(z_1W^{(\ell)}_1) = z_1\hbox{sign}(W^{(\ell)}_1).$$ Hence, in Algorithm~\ref{algo:correction} we obtain $g^{(\ell)}_1 = 1 > \varphi$. This means that the noisy node gets updated towards the correct direction.

Therefore, the only source of error would be a correct pattern neuron getting updated mistakenly. Let $P_{i}$ denote the probability that a correct pattern neuron $x^{(\ell)}_i$ gets updated. This happens if $\vert g^{(\ell)}_i \vert> \varphi$. For $\varphi \rightarrow 1$, this is equivalent to having $$\langle W^{(\ell)}_i,\hbox{sign}(z_1 W^{(\ell)}_1)\rangle = \Vert W^{(\ell)}_i \Vert_0.$$
However, in cases where the neighborhood of $x^{(\ell)}_i$ is different from the neighborhood of $x_1$ among the constraint nodes we have 
$$\langle W^{(\ell)}_i, \hbox{sign}(W^{(\ell)}_1) \rangle< \Vert W^{(\ell)}_i \Vert_0.$$
 More specifically, let $\mc{N}(x^{(\ell)}_i)$ indicate the set of neighbors of $x^{(\ell)}_i$ among constraint neurons in cluster $\ell$. Then in the case where $$\mc{N}(x^{(\ell)}_i) \cap \mc{N}(x^{(\ell)}_1) \neq \mc{N}(x^{(\ell)}_i), $$ there are non-zero entries in $W^{(\ell)}_i$ while $W^{(\ell)}_1$ is zero, and vice-versa. Therefore, by letting $P'_{i}$ to be the probability of $\mc{N}(x^{(\ell)}_i) \cap \mc{N}(x^{(\ell)}_1)= \mc{N}(x^{(\ell)}_i)$, we note that
\bed
P_{i} \leq P'_{i}.
\eed
The above inequality help us obtain an upper bound on $P_{i}$, by bound $P'_{i}$. Since there is only one noisy neuron $x_1$, we know that, on average, this node is connected to $\bar{d}_{\ell}$ constraint neurons which implies that the probability of $x_i$ and $x_1$ sharing exactly the same neighborhood is:
\bed
P'_{i} = \left(\frac{\bar{d}_{\ell}}{m_\ell} \right)^{d_{i}},
\eed
where $d_i$ is the degree of neuron $x_i$. By taking the average over the pattern neurons, we obtain the following bound on the average probability of a correct pattern neuron being mistakenly updated:
\beqa
P'_e &=& \sum_{d_i} \Lambda^{(\ell)}_{d_i} P'_{i} = \sum_{d_i} \Lambda^{(\ell)}_{d_i} \left(\frac{\bar{d}_{\ell}}{m_\ell} \right)^{d_{i}} = \Lambda^{(\ell)}\left(\frac{\bar{d}_{\ell}}{m_\ell}\right),\nonumber
\eeqa
where $\Lambda^{(\ell)}(x) = \sum_i \Lambda^{(\ell)}_i x^i$ is the degree distribution polynomial.
Therefore, the probability of correcting one noisy input is lower bounded by $P_c^{(\ell)} \geq \left(1-P'_e\right)^{n_\ell -1 }$, i.e.,
\bed
P_c^{(\ell)} \geq \left(1- \Lambda^{(\ell)}\left(\frac{\bar{d}_{\ell}}{m_\ell} \right) \right)^{n_\ell-1}.
\eed
This proves the theorem.

\subsection{Proof of Lemma~\ref{corrolary_single_error_no_two_column}}
Without loss of generality, suppose the first pattern neuron is contaminated by an external error $+1$, i.e., $z^{(\ell)} = [1,0,\dots,0]$. As a result
\bed
y^{(\ell)} = \hbox{sign}\left(W^{(\ell)}(x^{(\ell)}+z^{(\ell)})\right) = \hbox{sign}\left(Wx^{(\ell)} + Wz^{(\ell)}\right) = \hbox{sign}\left(W^{(\ell)}z^{(\ell)}\right) = \hbox{sign}\left(W^{(\ell)}_1\right),
\eed
where $W^{(\ell)}_i$ is the $\hbox{i}^{\hbox{th}}$ column of $W^{(\ell)}$. Hence, the feedback transmitted by the constraint neurons is $\hbox{sign}(W^{(\ell)}_1)$. As a result, decision parameters of pattern neuron $i$, i.e., $g^{(\ell)}_i$ in Algorithm \ref{algo:correction}, will be
\bed
g^{(\ell)}_i = \frac{\langle \hbox{sign}(W^{(\ell)}_1), W^{(\ell)}_i \rangle}{\langle \hbox{sign}(W^{(\ell)}_i), W^{(\ell)}_i \rangle}.
\eed
 Note that the denominator is simply $\Vert W^{(\ell)}_i \Vert_0 = \langle \hbox{sign}(W^{(\ell)}_i), W^{(\ell)}_i \rangle$. By assumption, no two pattern neurons in $G^{(\ell)}$ share the exact same set of neighbors. Therefore, for all $i,j \in\{ 1,\dots n_\ell\}$ such that $i\neq j$, there is at least a non-zero entry, say $k$, in $W^{(\ell)}_j$ for which $W^{(\ell)}_{ik} = 0$. Thus, we have $g^{(\ell)}_i =1$ if $i=1$ and $g^{(\ell)}_i <1$ if $i>1$
As a result, for $\varphi \rightarrow 1$, only the first neuron (i.e., the noisy one) will update its value towards the correct state.

\subsection{Proof of Theorem~\ref{th:outside}}

The proof is, in spirit, similar to Theorem 3.50 of \citep{urbanke}. Consider a message transmitted over an edge from a given cluster node $v^{(\ell)}$ to a given \emph{noisy} pattern neuron at iteration $t$ of Algorithm \ref{algo:peeling}. This message will be a failure, indicating that the super constraint node being unable to correct the error, if
\ben
\item the super constraint node $v^{(\ell)}$ receives at least one error message from its \emph{other} neighbors among pattern neurons. This event happens it is connected to more than one noisy pattern neuron.
\item the super constraint node $v^{(\ell)}$ does not receive an error message from any of its other neighbors but is unable to correct the single error in the given noisy neuron. This event happens with probability $1-P_c$.
\een 
 Let us denote the probability of the above failure message by $\pi^{(\ell)}(t)$ and the average probability that a pattern neuron sends an erroneous message to a neighboring cluster node by $z(t)$. 
%
Then, we have
\bed
\pi^{(\ell)}(t) = 1-P_c(1-z(t))^{\widetilde{d}_\ell-1},
\eed
where $\widetilde{d}_\ell$ is the degree of the super constraint neuron $v^{(\ell)}$ in the contracted graph $\widetilde{G}$. Similarly, let $\pi(t)$ denote the average probability that a super constraint node sends a message declaring the violation of at least one of its constraint neurons. Then we have
\bed
\pi(t) = \mb{E}_{\widetilde{d}_\ell} (\pi^{(\ell)}(t)) = \sum_{i} \widetilde{\rho}_i (1-P_c(1-z(t))^{\widetilde{d}_\ell-1}) = 1- P_c\widetilde{\rho}(1-z(t)).
\eed
Now consider the message transmitted from a given pattern neuron $x_i$ with degree $d_i$ to a given super constraint node $v^{(\ell)}$ in iteration $t+1$ of Algorithm \ref{algo:peeling}. This message will indicate a noisy pattern neuron if the pattern neuron was noisy in the first place (with probability $p_e$) and all of its \emph{other} neighbors among super constraint nodes has sent a violation message in iteration $t$. Therefore, the probability of this node being noisy will be $z(0) \pi(t)^{d_i-1}$ where $z(0) = p_e$. Hence, the average probability that a pattern neuron remains noisy at $(t+1)$-th iteration is 
\bed
z(t+1) = p_e \sum_i \widetilde{\lambda}_i \pi(t)^{i-1} = p_e \cdot \widetilde{\lambda}(\pi(t)) = p_e \cdot \widetilde{\lambda}(1- P_c\widetilde{\rho}(1-z(t))).
\eed
Note that the denoising operation will be successful if $z(t+1) < z(t),\ \forall t$. Therefore, we must look for the maximum $p_e$ such that $p_e \widetilde{\lambda}(1- P_c\widetilde{\rho}(1-z)) < z$ for $z \in [0,p_e]$.

\subsection{Proof of Theorem~\ref{theorem_exponential_solution}}
The proof is based on construction: we build a data set $\mc{X}$ with the required properties such that it can be memorized by the proposed neural network. 

Consider a matrix $G \in \mb{R}^{k \times n}$ with rank $k = rn$ where $0 < r < 1$ is chosen such that $k < \min_{\ell}(n_\ell)$. Let the entries of $G$ be non-negative integers between $0$ and $\gamma-1$. Here we assume that $\gamma \geq 2$. 

We start constructing the patterns in the data set as follows. We pick a random vector $u \in \mb{R}^k$ with integer-valued-entries between $0$ and $\upsilon-1$ where $\upsilon \geq 2$. We set the pattern $x \in \mc{X}$ to be $x = G^\top u$ \emph{if} all the entries of $x$ are between $0$ and $Q-1$. Since both $u$ and $G$ have only non-negative entries, all entries in $x$ are non-negative. However, we need to design $G$ such that all entries of $G^\top u$ be less than $Q$. Let $g_j$ be the $j$-th column of $G$. Then, the $j$-th entry of $x$ is equal to $x_j = \langle u, g_j\rangle$. Therefore,
\bed
x_j = u^\top g_j \leq d_j (\gamma-1) (\upsilon-1)
\eed
Let $d^* = \min_{j} d_j$. We can choose $\gamma$, $\upsilon$ and $d^*$ such that 
\bed
Q-1 \geq d^* (\gamma-1) (\upsilon-1)
\eed
which in turn ensures that all entries of $x$ are less than $Q$. Furthermore, we have selected $k$ in such a way that $k < \min_{\ell}(n_\ell)$. As a result we are sure that the set of sub-patterns of the dataset $\mc{X}$ form a subspace with dimension $k$ in an $n_\ell$-dimensional space. Since there are $\upsilon^k$ vectors $u$ with integer entries between $0$ and $\upsilon -1$, we have $\upsilon^k = \upsilon^{rn}$ patterns forming $\mc{X}$. This implies that the storage capacity $C = \upsilon^{rn}$ is an exponential number in $n$ as long as $\upsilon \geq 2$. 

%
%
%
%
%
%
%
%
%

\section{Conclusions and Final Remarks}\label{sec:convolutional_conclusions}
In this paper, we proposed the first neural network structure that learns an exponential number of patterns (in the size of the network) and corrects up to a linear fraction of errors. The main observation we made was that natural patterns seem to have inherent redundancy and we proposed a framework to captured redundancies that appear in the form of linear (or close to linear) constraints. Our experimental results also reveal that our learning algorithm can be seen as a feature extraction method, tailored for patterns with such constraints. Extending this line of thought through more sophisticated feature extraction approaches, and in light of recent developments in deep belief networks \citep{jarrett, coates, le, vincent,Ngiam}, is an interesting future direction to pursue.

%
%


\begin{small}
\bibliography{Neural_ref_with_url_doi}
\end{small}

\end{document}